\documentclass[11pt]{article}

\usepackage[]{acl}

\usepackage{times}
\usepackage{latexsym}

\usepackage[T1]{fontenc}

\usepackage[utf8]{inputenc}

\usepackage{microtype}

\usepackage{inconsolata}

\usepackage{graphicx}

\usepackage{bbold}
\usepackage{array}
\usepackage{mfirstuc}
\usepackage{subcaption}
\usepackage{verbatim}
\usepackage{fancyvrb}
\usepackage{adjustbox}
\usepackage{tabularx}
\usepackage{booktabs}

\usepackage{algpseudocode}
\usepackage{amsmath}
\usepackage{graphicx}
\usepackage{soul}
\sethlcolor{green!15}

\usepackage{tabularx}
\usepackage{array}
\usepackage{cellspace}
\usepackage{makecell}
\usepackage{amsmath}
\usepackage{adjustbox}  
\usepackage[most]{tcolorbox}
\usepackage{xspace}
\newcommand{\sysname}{\textsc{Minard}\xspace}
\usepackage[utf8]{inputenc}
\usepackage{tabularx}    
\usepackage{multirow}
\usepackage{microtype}
\usepackage{inconsolata}
\usepackage{inconsolata}
\usepackage{listings}
\usepackage[most]{tcolorbox} 
\usepackage{pifont}
\usepackage{multirow}
\usepackage{algorithm}
\usepackage{algpseudocode}
\usepackage{amsmath}

\usepackage[most]{tcolorbox}
\usepackage{enumitem}

\usepackage{booktabs}
\usepackage{array}
\usepackage{graphicx}
\usepackage{amssymb} 
\usepackage{booktabs}
\usepackage{subfiles} 
\usepackage[bb=ams]{mathalfa}

\title{Helping Figures Tell their Story! \\
Paper-Grounded Video Generation Explaining Complex Scientific Figures}

\definecolor{softgreen}{RGB}{235,255,235}

\definecolor{bertopiccolor}{HTML}{DD2829}  
\definecolor{ldacolor}{HTML}{3E7FB5}       
\definecolor{ctmcolor}{HTML}{5AAD50}       
\definecolor{dvaecolor}{HTML}{F98128}      

\newtcolorbox[list inside=prompt,auto counter,number within=section]{prompt}[1][]{
    colbacktitle=black!60,
    fonttitle=\small,
    coltitle=white,
    fontupper=\footnotesize,
    boxsep=3pt,
    boxrule=1pt,
    enhanced jigsaw,  
    #1,
}

\definecolor{best}{HTML}{5B6570}     
\definecolor{high}{HTML}{7D8894}     
\definecolor{mid}{HTML}{A2ACB5}
\definecolor{low}{HTML}{C6CDD3}
\definecolor{weak}{HTML}{E7EBEE}

\newcommand{\bestcell}[1]{\cellcolor{best}\textcolor{white}{#1}}
\newcommand{\highcell}[1]{\cellcolor{high}#1}
\newcommand{\midcell}[1]{\cellcolor{mid}#1}
\newcommand{\lowcell}[1]{\cellcolor{low}#1}
\newcommand{\weakcell}[1]{\cellcolor{weak}#1}

\usepackage{array}

\usepackage{subcaption}

\usepackage{booktabs}

\usepackage{algorithm}

\usepackage{algpseudocode}
\newif\ifcomment\commenttrue

\usepackage[a-1b]{pdfx}

\usepackage{framed}
\usepackage{mdwlist}
\usepackage{siunitx}
\usepackage{latexsym}
\usepackage{colortbl}

\usepackage{nicefrac}
\usepackage{booktabs}
\usepackage{fnpct}
\usepackage{amsfonts}
\usepackage[T1]{fontenc}
\usepackage{bold-extra}
\usepackage{amsmath}
\usepackage{amssymb}
\usepackage{bm}
\usepackage{graphicx}
\usepackage{mathtools}
\usepackage{microtype}
\usepackage{centernot}
\usepackage{multicol}
\usepackage{xpatch}
\usepackage[inkscapelatex=false]{svg}
\usepackage{latexsym,comment}
\usepackage[normalem]{ulem}

\newcommand*{\missingreference}{{\Huge \colorbox{red}{?reference?}}}
\newcommand*{\missingcitation}{{\Huge \colorbox{red}{?citation?}}}

\makeatletter
\xpatchcmd{\@setref}{\bfseries}{\missingreference}{}{}
\def\@citex[#1]#2{\leavevmode
    \let\@citea\@empty
    \@cite{\@for\@citeb:=#2\do
        {\@citea\def\@citea{,\penalty\@m\ }%
            \edef\@citeb{\expandafter\@firstofone\@citeb\@empty}%
            \if@filesw\immediate\write\@auxout{\string\citation{\@citeb}}\fi
            \@ifundefined{b@\@citeb}{\hbox{\reset@font\missingcitation}%
                \G@refundefinedtrue
                \@latex@warning
                {Citation `\@citeb' on page \thepage \space undefined}}%
            {\@cite@ofmt{\csname b@\@citeb\endcsname}}}}{#1}}
\makeatother

\newcommand{\gem}[1]{\mbox{\textsc{gem}}}

\usepackage{amsmath,amsfonts,bm}

\def\eqref#1{equation~\ref{#1}}

\def\1{\bm{1}}

\DeclareMathAlphabet{\mathsfit}{\encodingdefault}{\sfdefault}{m}{sl}
\SetMathAlphabet{\mathsfit}{bold}{\encodingdefault}{\sfdefault}{bx}{n}

\renewenvironment{quote}
{\list{}{\rightmargin\leftmargin}%
    \item\relax\small\ignorespaces}
{\unskip\unskip\endlist}

\newcommand{\hidetext}[1]{}
\newcommand{\ignore}[1]{}
\newif\ifcomment
\commenttrue

\ifcomment
    \newcommand{\pinaforecomment}[3]{\colorbox{#1}{\parbox{.8\linewidth}{#2: #3}}}

    \newcommand{\prtodo}[1]{\pinaforecomment{lightblue}{pr}{#1}}
    \newcommand{\prtodoi}[1]{\pinaforecomment{lightblue}{pr}{#1}}
\else
    \newcommand{\pinaforecomment}[3]{}
    \newcommand{\prtodo}[1]{}
    \newcommand{\prtodoi}[1]{}
\fi

\usepackage{colortbl}
\definecolor{best}{HTML}{1B5E20}     
\definecolor{good}{HTML}{66BB6A}     
\definecolor{mid}{HTML}{FFEB3B}      
\definecolor{weak}{HTML}{FFA726}     
\definecolor{worst}{HTML}{E53935}    
\newcommand{\cbest}[1]{\cellcolor{best!55}\textbf{#1}}
\newcommand{\cgood}[1]{\cellcolor{good!40}#1}
\newcommand{\cmid}[1]{\cellcolor{mid!35}#1}
\newcommand{\cweak}[1]{\cellcolor{weak!35}#1}
\newcommand{\cworst}[1]{\cellcolor{worst!30}#1}

\newcommand{\smallurl}[1]{ \begin{tiny}\url{#1}\end{tiny}}

\definecolor{lightblue}{HTML}{3cc7ea}
\definecolor{CUgold}{HTML}{CFB87C}
\definecolor{grey}{rgb}{0.95,0.95,0.95}
\definecolor{ceil}{rgb}{0.57, 0.63, 0.81}
\definecolor{UMDred}{HTML}{ed1c24}
\definecolor{UMDyellow}{HTML}{ffc20e}

\definecolor{shadeA}{HTML}{8FC6CB}  
\definecolor{shadeB}{HTML}{B0D6D9}
\definecolor{shadeC}{HTML}{CBE3E5}
\definecolor{shadeD}{HTML}{E0EFF0}
\definecolor{shadeE}{HTML}{F0F7F8}  

\usepackage{xspace}
\newcommand{\eg}{e.g.\xspace}

\usepackage{pifont}

\usepackage{xcolor}

\definecolor{promptbg}{HTML}{F6F7F9}
\definecolor{promptborder}{HTML}{B8C0C8}
\definecolor{prompttitle}{HTML}{4A5560}

\lstdefinestyle{promptstyle}{
    basicstyle=\ttfamily\tiny,
    breaklines=true,
    columns=fullflexible,
    keepspaces=true,
    frame=none,
    showstringspaces=false
}

\newcommand{\cmark}{\textcolor{green!70!black}{\ding{51}}}
\newcommand{\xmark}{\textcolor{red!80!black}{\ding{55}}}

\author{$^{*}$Ishani Mondal,    
 \thanks{Equal Contribution} \textbf{Javad Baghirov}, \textbf{Jordan Boyd-Graber} \\ \\
   University of Maryland, College Park \hspace{0.1cm}
}

\begin{document}
\maketitle
\begin{abstract}
Scientific figures compress complex pipelines into a single canvas, yet understanding them requires paper-grounded, step-by-step narration aligned with visual highlights—a capability missing from current video generation systems and benchmarks \citep{Mayer_2009, sweller1988cognitive, hsu-etal-2021-scicap-generating, masry-etal-2022-chartqa, EbrahimiKahou2017FigureQAAA}.
To address this, we introduce \textbf{paper-grounded figure-to-video generation}: generating narrated, region-grounded walkthrough videos from a figure and its paper. We propose \textsc{\sysname} (\textbf{M}ultimodal \textbf{I}nterpretation of \textbf{N}arrated \textbf{A}rchitecture via \textbf{R}egion \textbf{D}ecomposition), a pipeline that generates paper-grounded narrations and sequentially grounds them to figure regions. We also release \textbf{FigTalk}, a benchmark with new sequential and component-level grounding metrics derived. 
On FigTalk, \textsc{\sysname} generates human-like, paper-faithful narrations and outperforms narration-conditioned figure spatial grounding compared to existing approaches in both automatic and human evaluation.
\end{abstract}

\section{From Static Figure Summary to Narrated Step-by-Step Explanations}
We have all been there. A conference talk flashes an architecture diagram for a few seconds, you scramble to follow the arrows and identify the novelty, and before it clicks the speaker says ``the figure speaks for itself'' and advances the slide. Good speakers do the opposite: they walk the audience through the setup (``input on the left, output on the right''), the semantics (``boxes are modules, arrows are data flow''), what matters (``the orange block is the important bit: a new Fourier transform''), and the takeaway (``the change in representation helps the compression happen''). 
Multimedia learning theory explains why such walkthroughs help: comprehension improves when narration is temporally aligned with visual regions (\emph{temporal contiguity}), attention is guided to salient components (\emph{signaling}), and explanations are segmented into steps \citep{Mayer_2009, sweller1988cognitive, MOOC}.
A single diagram can convey what would take pages of prose~\citep{LARKIN198765}, but that density is exactly what makes figures cognitively demanding, forcing readers to integrate labels, arrows, modules, and visual structure with the surrounding paper \citep{Hegarty} which is why authors routinely explain their own figures step-by-step in talks to make them more comprehensible.

\begin{figure}[t]
  \centering
  \includegraphics[width=0.50\textwidth]{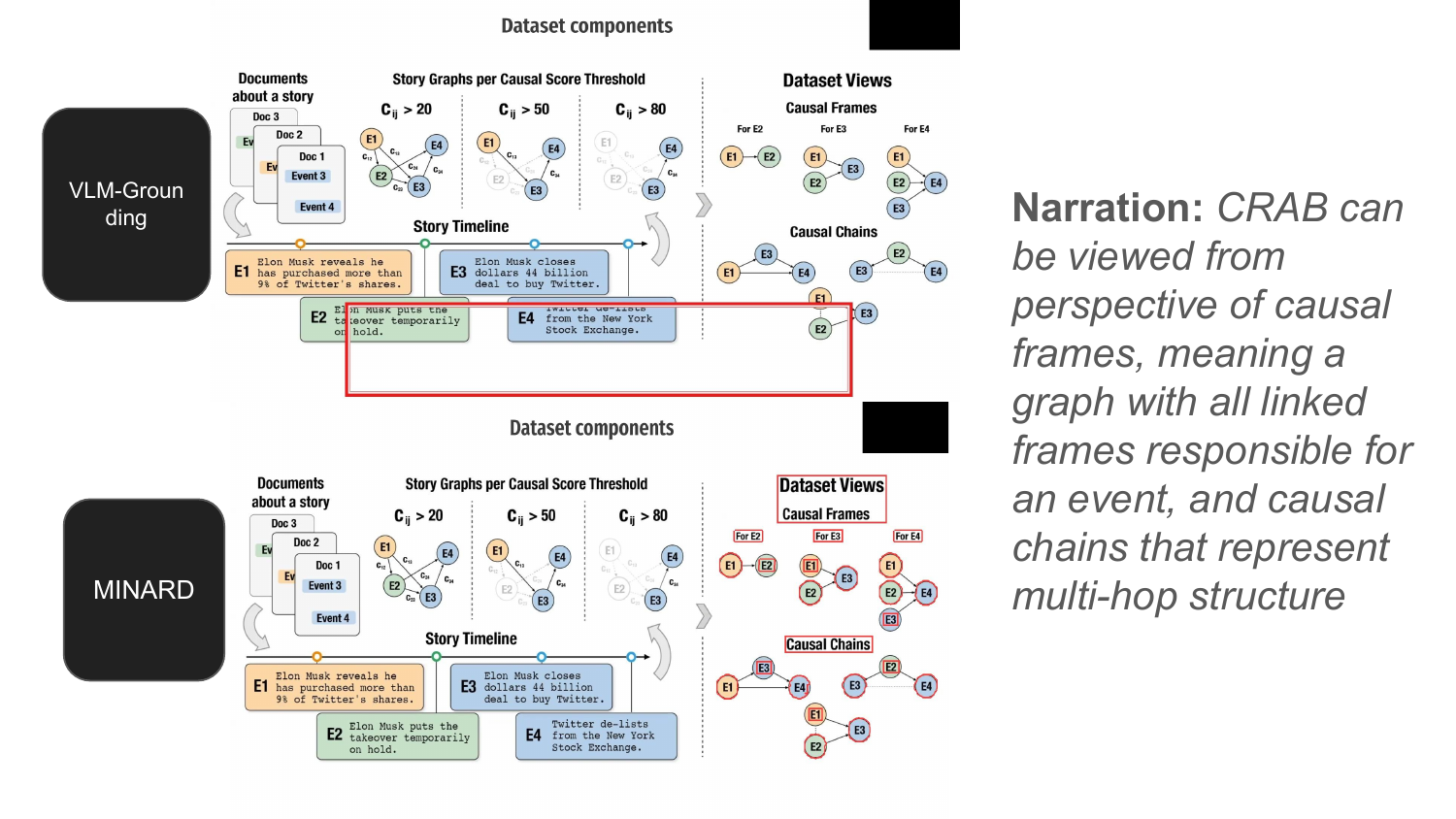}
  \caption{On a structurally complex figure with long-range causal dependencies, \sysname{} produces grounding patterns closely aligned with human explanations, while direct VLM-grounding fail to preserve the correct causal structure, temporal ordering, and cross-component associations (Gold and Veo-3.1 in ~\ref{fig:example_input_hallu}).}
  \label{fig:example_input1}
  \vspace{-2.25mm}
\end{figure}

Can AI generate the same kind of explanation, and how would we know if it succeeded? Existing approaches do not answer either question because they treat figures as static images rather than structured, narrated explanations. Figure captioning describes \emph{what} appears in an image rather than \emph{why} a system works \citep{hsu-etal-2021-scicap-generating}; chart and figure QA focus on isolated lookups instead of coherent exposition \citep{masry-etal-2022-chartqa, EbrahimiKahou2017FigureQAAA}; and visual grounding localizes single queries without constructing a step-by-step walkthrough.
Existing systems also fall short. TheoremExplainAgent \citep{ku-etal-2025-theoremexplainagent} and Code2Video \citep{code2video} generate animated explanations from abstract concepts rather than grounding to a source figure \citep{ku2025theoremexplainagentvideobasedmultimodalexplanations}. Presentation-generation systems such as Doc2PPT \citep{fu2022doc2pptautomaticpresentationslides}, D2S \citep{sun-etal-2021-d2s}, PPTAgent \citep{zheng-etal-2025-pptagent}, and Paper2Video \citep{zhu2025paper2videoautomaticvideogeneration} treat figure explanation as a small part of presentation generation, typically moving a cursor across regions without reasoning about what to explain, in what order, or why components matter. 
Diffusion-based video generators such as Veo \citep{Yang2024CogVideoXTD} hallucinate unsupported details (Figure~\ref{fig:example_input1}). Across all these approaches, figures are treated as opaque images rather than structured sequences of explainable components, and no benchmark evaluates sequential, paper-grounded figure explanation.

To address these limitations, we introduce \emph{paper-grounded figure explanation}, where a system generates a step-by-step narration of a paper figure with aligned visual highlights. We present \textbf{FigTalk}, the first benchmark for narrated architectural-figure walkthroughs derived from conference presentations, together with evaluation protocols for sequential narration and grounding (Section~\ref{sec:figtalk}). We further introduce \textsc{\sysname}\footnote{In honor of Charles Joseph Minard, whom Edward Tufte immortalized as the master of explicating complex ideas visualized \citep{tufte1983visual}}, a pipeline that decomposes figures into structured components, generates paper-grounded narration, and sequentially aligns narration steps to figure regions (Section~\ref{sec:figlecturer}). On FigTalk, \sysname substantially improves narration faithfulness and takeaway quality, achieves the best grounding performance, and is preferred by humans in 74\% of narration comparisons and 63\% of grounding comparisons, especially on harder figures (Sections~\ref{sec:results} and~\ref{sec:human-systems}).

\section{FigTalk: Benchmarking Narrated Figure Explanations}
\label{sec:figtalk}

\begin{table}[t]
\centering
\small
\begin{tabular}{lcccc}
\toprule
 & Easy & Medium & Hard & Total \\
\midrule
\textbf{FigTalk-Gold}    & 18 & 22 & 9  & 49 \\
\textbf{FigTalk-Extended}  & 20 & 30 & 15 & 65 \\
\midrule
Total                      & 38 & 52 & 24 & 114 \\
\bottomrule
\end{tabular}
\caption{\textsc{FigTalk} composition. The reference set ($n{=}49$) carries ground-truth grounding traces; the extended set ($n{=}65$) is scored via the rubric in \S\ref{sec:eval-grounding}.}
\label{tab:figtalk-stats}
\end{table}

To test whether a model can \emph{narrate} a figure---trace its components as a coherent walkthrough over time---we focus on \emph{architectural figures}, where understanding depends on explaining interactions between modules, data flow, and reasoning steps. Prior systems such as TheoremExplainAgent \citep{ku-etal-2025-theoremexplainagent} and Code2Video \citep{code2video} already generate Manim-style animations for charts, equations, and abstract concepts, but they do not explain complex architectural figures grounded in a source paper. Evaluating this capability therefore requires a benchmark of videos grounded in individual architectural figures.
%
This section describes how we build one. 

\paragraph{Sources.} 
We construct FigTalk from paper--video pairs in VISTA~\citep{liu-etal-2025-talk} and Paper2Video~\citep{zhu2025paper2videoautomaticvideogeneration}, restricting to \emph{architectural and method-description figures}---pipelines, system diagrams, and schematic overviews---where understanding depends on sequential explanation of interacting components. We exclude charts and result plots, since recent systems such as TheoremExplainAgent already study animated explanations for those settings~\citep{ku2025theoremexplainagentvideobasedmultimodalexplanations}.
Two authors independently review each candidate video to verify that (i) a paper figure is visibly presented, (ii) the speaker explicitly walks through the figure step by step, and (iii) the explanation remains faithful to the paper and visually grounded to the discussed regions. The \textbf{49 videos} satisfying all criteria (Cohen's $\kappa{=}0.81$) form \textbf{FigTalk-Gold}, our fully verified reference subset with exact narration--grounding traces. Each example contains: (1) the paper and target figure, (2) ordered narration segments ${s_i^*}{i=1}^{T}$, and (3) synchronized grounding traces ${G_i}{i=1}^{T}$ specifying the figure regions emphasized at each narration step.
For each FigTalk-Gold video, we identify the figure-explanation segment, extract frames at 2 FPS, blur frames to suppress minor noise, and detect visual changes between adjacent frames using SSIM. Peaks in change typically correspond to cursor movements, zooms, annotations, or shifts in presenter focus, defining candidate grounding steps. For each step, we extract the aligned audio segment to obtain narration units $s_i^\ast$ and record the newly emphasized figure regions $G_i$, yielding synchronized narration--grounding traces $\{(s_i^\ast, G_i)\}_{i=1}^{T}$. Two authors independently verify all step boundaries and region assignments, resolving disagreements through discussion.
To enable broader evaluation, we additionally collect \textbf{FigTalk-Extended}, a set of \textbf{65 videos} where both annotators agree that an architectural figure is being explained, but grounding traces are weaker or incomplete. Each example contains the paper, target figure, and narration transcript, but only coarse or partial grounding supervision. These examples are evaluated using the rubric-based protocols of \S\ref{sec:eval-grounding} rather than exact grounding alignment. The final benchmark contains 114 figure--video pairs (Table~\ref{tab:figtalk-stats}).

\textbf{Complexity stratification.} Explanation difficulty grows with a figure's internal structure, not its size, so we stratify on three automatically extracted signals---component count, DAG depth, and sub-figure count---into Easy/Medium/Hard tiers. 
The signals are extracted by a vision--language model (Gemini~3 Pro) prompted to count high-level semantic units rather than low-level visual primitives, and a fixed deterministic rule maps the counts to a tier (Details in App.~\ref{app:complexity}). This lets us report whether a system's gap widens on Hard figures, the regime where animation should help most.

\begin{figure*}[t]
  \centering
    \fbox{\includegraphics[width=0.99\textwidth]{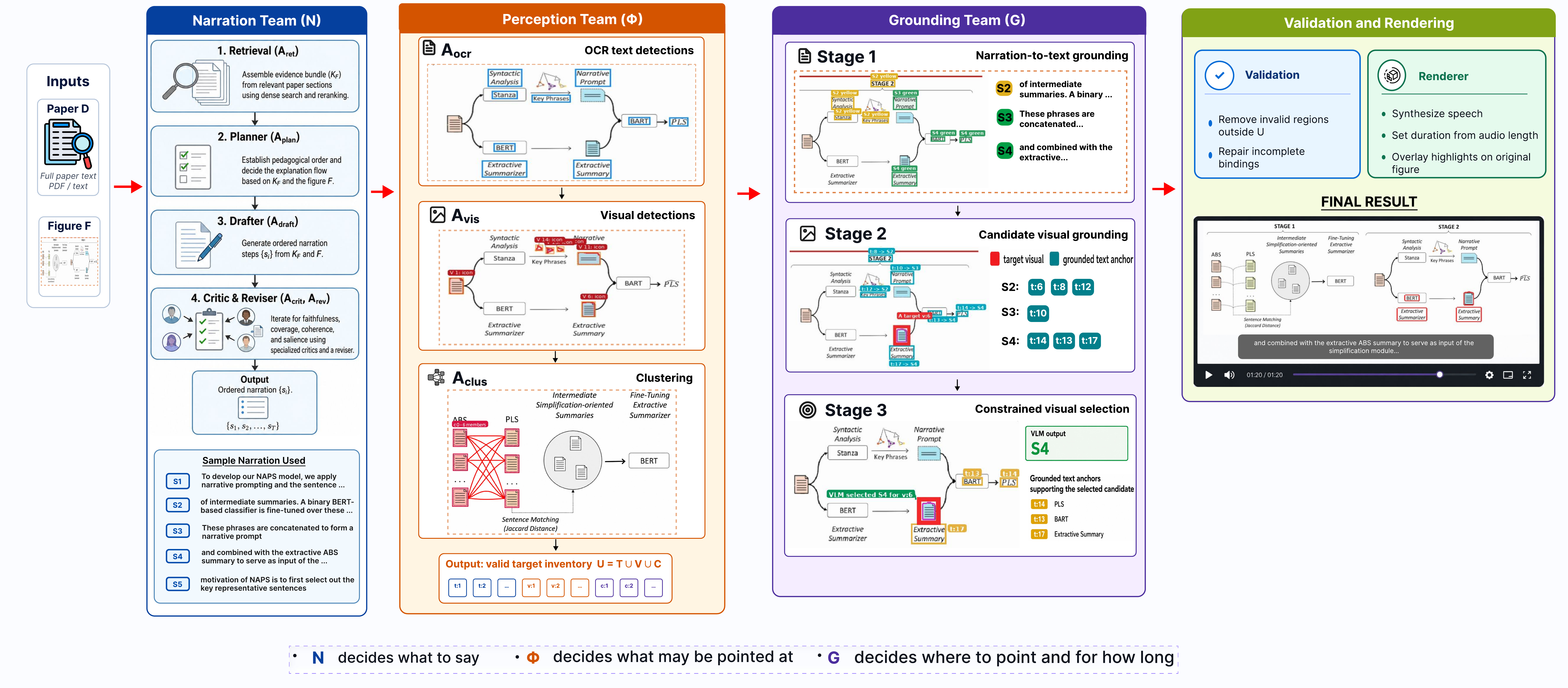}}
  \caption{Overview of \sysname{} on the NapSS~\cite{lu-etal-2023-napss} architecture, illustrating how narration generation, figure perception, and constrained grounding collaborate to produce figure-grounded explanatory videos.}
  \label{fig:example_input}
\end{figure*}

\section{\sysname}
\label{sec:figlecturer}


\sysname{} turns a paper $D$ and a target figure $F$ into a narrated, figure-grounded video $V$ through three specialized teams: the \textbf{Narration team} $\mathcal{N}$ reads the paper and decides \emph{what should be said}, producing an ordered narration $\{s_i\}_{i=1}^{T}$, the \textbf{Perception team} $\Phi$ reads the figure and decides \emph{what may be pointed at}, constructing a pool of valid visual targets $\mathcal{U}$ from OCR, detection, segmentation, and layout cues and the \textbf{Grounding team} $\mathcal{G}$ then decides \emph{where to look and for how long}, linking each narration step $s_i$ to regions in $\mathcal{U}$ and rendering the final highlighted video.
We use the NapSS figure of Fig.~\ref{fig:example_input1} from \cite{lu-etal-2023-napss} as a running example throughout: a two-stage summarization pipeline whose Stage~1 builds intermediate simplification-oriented summaries, and whose Stage~2 runs a syntactic-analysis branch, merging both through BART to produce the final PLS.

\subsection{Narration Team ($\mathcal{N}$)}
\label{sec:narration}
$\mathcal{N}$ is the only team that reads the full paper and it chains one retrieval step with four LLM agents to produce the contract $\{s_i\}$:

\noindent\textbf{Retrieval $\mathcal{A}_{\mathrm{ret}}$}: The NapSS figure could be read many ways---as a list of model blocks (BERT, Stanza, BART), as the flow of data from ABS/PLS to the final PLS, or as the paper's core idea of \emph{narrative prompting}---and which reading the authors intend is not recoverable from $F$ alone. So $\mathcal{A}_{\mathrm{ret}}$ uses dense embedding search plus cross-encoder reranking---not generation---to assemble a figure-specific evidence bundle $K_F$: the caption, every in-text reference to the figure, the paragraphs describing Stage~1 and Stage~2, and the contribution sentences that frame narrative prompting as the key novelty. This grounds every later decision in the paper's own text rather than LLM priors (App.~\ref{app:retrieval}).

\noindent\textbf{Planner \& Drafter
$\mathcal{A}_{{\mathrm{plan}}},\mathcal{A}_{{\mathrm{draft}}}$}:
A good explanation follows a teaching order rather than a random traversal of figure components. Conditioned on $K_F$ and $F$, the planner LLM recovers a pedagogical sequence of concepts and supporting claims, while the drafter LLM converts them into an ordered narration ${s_i}$ aligned with the figure structure. For the NapSS example, this produces a six-step walkthrough moving from inputs and preprocessing, through the parallel summarization branches, to fusion and final summary generation. Full prompts are provided in Appendix~\ref{app:prompts}.

\noindent\textbf{Critic \& Reviser
$\mathcal{A}_{\mathrm{crit}}^{1:4},\mathcal{A}_{\mathrm{rev}}$:}
A one-pass draft tends to drift, so four critic LLMs each check one failure mode against $K_F$: \textit{faithfulness} (the draft must not claim BART is fine-tuned in Stage~1 when only BERT is); \textit{coverage} (the narrative-prompt branch must not be dropped in favor of only the extractive branch); \textit{coherence} (steps must follow the left-to-right, stage-1-then-stage-2 traversal without jumping back); and \textit{salience} (merging redundant mentions of the document icons so the prompt-fusion novelty stays foregrounded). A reviser LLM merges these critiques into a revised $\{s_i\}$; the loop stops when critics pass or a budget is hit, and $\{s_i\}$ is written \textsc{final} (App.~\ref{app:critics}).

\subsection{Perception Team ($\Phi$)}
\label{sec:extraction}
$\Phi$ decides what parts of $F$ can actually be highlighted, so later stages point only at real elements instead of hallucinating locations. It relies on perception models and post-processing rather than free-form generation, and sees only $F$, never $\{s_i\}$, keeping regions independent of explanation.

\noindent\textbf{Text \& Vision
$\mathcal{A}_{\mathrm{ocr}},\mathcal{A}_{\mathrm{vis}}$.}
Figures hold both text and visuals, so two models run in parallel. On NapSS, an OCR model extracts text regions $\mathcal{T}$ such as the labels \textsc{abs}, \textsc{pls}, \emph{Sentence-Matching (Jaccard Distance)} etc; open-vocabulary detection and segmentation models extract visual regions $\mathcal{V}$ such as the document icons, the Stage~1 blob, the three model boxes, the key-phrases glyph, and the connecting arrows (App.~\ref{app:detect}).


\noindent\textbf{Cluster $\mathcal{A}_{\mathrm{clus}}$.}
A label and its diagram block are only meaningful together, so $\mathcal{A}_{\mathrm{clus}}$ uses visual embeddings to group detections that are visually close to each other --- for instance binding the 6 document icons into a cluster $\mathcal{C}$ that can later be highlighted as a single unit.

\noindent\textbf{Builder $\mathcal{A}_{\mathrm{reg}}$.}
Final step assigns stable IDs and bounding boxes and outputs the whitelist of valid targets $\mathcal{U} = \mathcal{T} \cup \mathcal{V} \cup \mathcal{C}$, which depends only on $F$ (App.~\ref{app:builder}). Optionally it tags coarse spatial zones (the Stage~1 left region versus the Stage~2 right region) to disambiguate the visually similar document icons that appear in both halves.

\subsection{Grounding Team ($\mathcal{G}$)}
\label{sec:grounding} 
$G$ aligns each narration step $s_i$ with the correct regions in the figure. Unlike unconstrained VLM grounding, \sysname performs grounding as a staged, semantically constrained selection process over the validated region inventory $U$ produced by $\Phi$. This prevents the system from hallucinating regions that do not exist in the original figure.

\paragraph{Stage 1: Narration-to-text grounding.}
Figure explanations are typically anchored around textual semantics such as module names and operation headers. \sysname therefore first grounds narration steps to OCR-derived text regions before grounding visual components. Given OCR regions $T=\{t_1,\dots,t_n\}$ and narration steps $\{s_i\}_{i=1}^{T}$, a language model predicts mappings of the form $s_i \rightarrow \{t_j\}$. For the NapSS example, the narrative-prompting step grounds to regions such as \texttt{STANZA}, \texttt{Key Phrases}, and \texttt{Narrative Prompt}.

\paragraph{Stage 2: Candidate visual grounding.}
The grounded text regions are then used to constrain visual grounding. For each visual region $v \in V$, \sysname identifies nearby grounded text regions using spatial proximity and cluster membership, and inherits their associated narration steps as candidate narration steps, producing mappings $v \rightarrow \mathcal{S}_v$. This converts grounding from a global search into a localized semantically conditioned matching problem. For example, the STANZA module inherits candidate narration steps from nearby labels such as \texttt{STANZA} and \texttt{Key Phrases}.

\paragraph{Stage 3: Constrained visual selection.}
\sysname performs constrained visual grounding independently for each visual region. A vision-language model receives: (i) the figure with one candidate visual region highlighted, and (ii) only candidate narration steps $\mathcal{S}_v$ associated with that region. The model predicts which narration steps should ground to highlighted region. Restricting the model to semantically filtered candidates avoids the broad over-highlighting behavior common in unconstrained VLM grounding. In the NapSS example, the narrative-prompt branch jointly highlights the STANZA module, key-phrase extraction block, and Narrative Prompt component while excluding the unrelated BERT extractive branch.

\paragraph{Validation and rendering.}
Finally, a deterministic validator removes invalid bindings outside the whitelist $U$ and repairs incomplete selections. The renderer synthesizes the narrated walkthrough video by aligning each grounded region sequence with narration timing, preserving temporal alignment between speech and visual focus (grounded-script format and rendering details in App.~\ref{app:grounding}).

\medskip
Overall, \sysname{} splits figure explanation into three clear responsibilities---\emph{what to say} ($\mathcal{N}$), \emph{what is valid to point at} ($\Phi$), and \emph{where to point} ($\mathcal{G}$). Learned LLM decisions are confined to narration authoring and region selection, while perception and rendering rely on dedicated models and deterministic steps, making the system reliable, interpretable, and easy to evaluate.

\section{Experimental Setup and Evaluation}
\label{sec:experiments}
\textsc{\sysname} separates figure explanation into two stages: a narrator generates explanation steps $\{s_i\}_{i=1}^{T}$, and a grounder aligns each step to figure regions, producing $\{(s_i, R_i, t_i)\}_{i=1}^{T}$. Since failures may arise from either stage, we study them separately through three different research questions: \textbf{RQ1)} Does conditioning narration on the paper improve explanation quality? \textbf{RQ2)} Does explicit figure decomposition improve narration-to-region grounding over existing baselines? \textbf{RQ3)} Which components of \textsc{\sysname} contribute most to these gains?
Accordingly, we evaluate (1) \emph{narration quality} ($D_1$), comparing generated $\{s_i\}_{i=1}^{T}$ against human narration $\{s_i^*\}_{i=1}^{T}$ from FigTalk, and (2) \emph{grounding quality conditioned on narration} ($D_2$), comparing predicted grounding traces $\{(R_i, t_i)\}_{i=1}^{T}$ against human grounding traces on FigTalk-Gold, alongside non-reference grounding metrics on FigTalk-Extended while fixing narration to $\{s_i^*\}_{i=1}^{T}$. RQ3 is studied through ablations.
\subsection{Narration Evaluation (RQ1)}
\label{sec:eval-narration}
\paragraph{Baselines.}
To isolate the effect of narration conditioning, we vary only the narrator input while keeping the rest of the pipeline fixed: \textbf{(a) Figure-grounded}, using only the figure $F$; \textbf{(b) Paper-and-figure-grounded}, using both the figure $F$ and paper $D$ (default \textsc{\sysname} Narration Team); and \textbf{(c) Paper2Video SlideTalker}~\citep{zhu2025paper2videoautomaticvideogeneration}, which narrates around embedded figures rather than through their components. Each setting is evaluated with three frontier multimodal backbones (\textsc{Gemini-3.1-Pro}, \textsc{GPT-5}, and \textsc{Claude Sonnet}); implementation details appear in Appendix~\ref{app:narration-baselines}.

\paragraph{Metrics.}
Our narration evaluation framework treats the paper and figure as the only sources of factual truth. The gold narration $Y^\star$ is used only to evaluate \textbf{Order Matching}, \textbf{Concept Coverage}, and \textbf{Takeaway Recall} using the procedures in Appendix~\ref{app:narration-metrics}. It is not treated as an oracle for factual correctness, since multiple faithful explanations may exist for the same figure. Instead, \textbf{Internal Faithfulness} measures whether narration claims are supported by the paper and figure evidence bundle $\mathcal{K}_F$, while \textbf{External Faithfulness} verifies added background information against external references $\mathcal{K}_{\mathrm{ext}}$. This design avoids penalizing explanations that differ from a presenter’s narration but remain factually grounded.

\subsection{Grounding on Narration (RQ2)}
\label{sec:eval-grounding}
We hold the narration $\{s^*_i\}_{i=1}^{T}$ and vary only the grounding pipeline; every system emits a timed region trace $\{(s^*_i, R_i, t_i)\}_{i=1}^{T}$ over $F$. This isolates grounding capability from narration capability and makes $D_2$ directly comparable across families that would otherwise produce incompatible narrations.

\paragraph{Baselines.}
We compare \textsc{\sysname} against four families that span the design space of narration-to-region binding using the Figure $F$ and human-authored narration $\{s^*_i\}_{i=1}^{T}$  as inputs.
\textbf{(i) VLM-Grounding} bypasses the entire Perception Team $\Phi$ in \sysname and directly replaces the Grounding Team $G$ with a single unconstrained VLM prediction step, given the Figure $F$ and the image width and height, and the narration steps. 
\textbf{(ii) SAM Segmentation + BBox} partially overlaps with the Perception Team $\Phi$, using OCR and SAM-based extraction to generate candidate regions, but omits the structured processing that makes $\Phi$ semantically meaningful. Its grounding stage is simplified than \sysname since the LLM directly selects candidate region IDs from the raw extraction list. Both (i) and (ii) are implemented using three model backbones as specified in the Narration Generation Stage (D1).
\textbf{(iii) Manim-Code Adaptations}  such as Code2Video \citep{code2video}  and \textsc{TheoremExplainAgent}~\citep{ku-etal-2025-theoremexplainagent}, which we adapt using $F$ and narration as inputs; 
and \textbf{(iv) Cursor-Grounded Adaptation using SlideTalker} based on Paper2Video \citep{zhu2025paper2videoautomaticvideogeneration} 
and \textbf{(v) Diffusion Video Generation} such as Veo 3.1~\citep{wiedemer2025videomodelszeroshotlearners} and CogVideoX~\citep{Yang2024CogVideoXTD} where we use the text-to-video diffusion conditioned on $\{s^*_i\}$ and the figure. 
All the implementation details are in Appendix~\ref{app:grounding-baselines}.




\paragraph{Metrics.} We evaluate grounding under two complementary protocols depending on whether gold grounding traces are available. For the \textbf{FigTalk-Gold}, we construct ordered gold grounding sequences from human conference videos and align system predictions to these sequences using Dynamic Time Warping (DTW) \citep{senin2008dynamic}. 
We report two DTW-based metrics: \textbf{soft-DTW}, a frame-level similarity metric that uses a calibrated VLM judge and applies to all systems including video-generation models, and \textbf{IoU-based DTW}, a stricter region-level metric defined only for systems that output explicit grounding regions. IoU-based DTW computes per-step precision, recall, and F1 from matched predicted and gold regions and reports \textbf{Macro-P}, \textbf{Macro-R}, and \textbf{Macro-F1} after DTW alignment, while soft-DTW measures frame-level grounding similarity through judge-scored alignment costs.
We use \textbf{soft-DTW} to evaluate all systems uniformly from rendered frames, including video-generation models without explicit grounding outputs, and \textbf{IoU-based DTW} to provide stricter region-level precision, recall, and F1 evaluation for systems that predict explicit grounding regions.
For the \textbf{FigTalk-Extended}, we use a rubric-based evaluation along four explanatory axes: \textbf{Inputs}, \textbf{Mechanism}, \textbf{Outputs}, and \textbf{Takeaway}. Each axis receives three judgments: \textbf{Component Faithfulness (CF)} (whether highlighted regions and narration are scientifically correct), \textbf{Concept Coverage (CC)} (whether important concepts are both highlighted and explained), and \textbf{Excess Highlight Rate (EH)} (whether the system over-highlights irrelevant or unexplained regions). 
We additionally validate the VLM-based evaluation protocol against human annotators and report inter-annotator agreement as well as VLM--human agreement. Full metric definitions and rubric instructions are provided in Appendix~\ref{app:grounding-metrics}.

\subsection{Ablation Study Setup (RQ3)}
\label{sec:eval-ablation}
\label{sec:eval-human}
For RQ3 we run two ablation families. \emph{Narration-side} ablations (Table~\ref{tab:narration_ablation2}) toggle retrieval ($K_F$), the critic--reviser loop, the ordering planner, and takeaway refinement. \emph{Grounding-side} ablations (Table~\ref{tab:grounding_ablation}) toggle the components of \textsc{\sysname} Perception module components. 


\begin{table*}[t]
\centering
\setlength{\tabcolsep}{3pt}
\begin{tabular}{llccccc}
\toprule
Regime & Backbone & Order & Internal & Concept & Takeaway & External \\
       &          & Match. & Faith.  & Cov.    & Recall   & Faith.   \\
\midrule
\multirow{3}{*}{\shortstack[l]{Figure-only\\(F only)}}
& \textsc{Gemini} & \cmid{0.71} & \cweak{0.78} & \cweak{0.56} & \cweak{0.39} & \cworst{0.24} \\
& \textsc{Claude}  & \cmid{0.72} & \cmid{0.79} & \cweak{0.58} & \cweak{0.41} & \cworst{0.26} \\
& \textsc{GPT-5}          & \cmid{0.73} & \cmid{0.80} & \cmid{0.60} & \cweak{0.43} & \cworst{0.28} \\
\midrule
\multirow{3}{*}{\shortstack[l]{Paper2Video \\ \citep{zhu2025paper2videoautomaticvideogeneration} \\ (Figure + Document)}}
& \textsc{Gemini} & \cworst{0.66} & \cweak{0.77} & \cworst{0.54} & \cworst{0.31} & \cweak{0.39} \\
& \textsc{Claude}  & \cworst{0.68} & \cmid{0.79} & \cworst{0.57} & \cworst{0.33} & \cweak{0.42} \\
& \textsc{GPT-5}          & \cweak{0.69} & \cmid{0.81} & \cweak{0.59} & \cworst{0.36} & \cweak{0.45} \\
\midrule
\multirow{3}{*}{\shortstack[l]{\sysname\\(Figure + Document)}}
& \textsc{Gemini} & \cgood{0.74} & \cmid{0.80} & \cgood{0.75} & \cgood{0.70} & \cgood{0.74} \\
& \textsc{Claude}  & \cgood{0.75} & \cgood{0.82} & \cgood{0.79} & \cgood{0.76} & \cgood{0.78} \\
& \textsc{GPT-5}          & \cbest{0.76} & \cbest{0.83} & \cbest{0.84} & \cbest{0.81} & \cbest{0.82} \\
\bottomrule
\end{tabular}
\caption{$D_1$ narration quality on FigTalk-Gold. Color encodes within-column rank: \textcolor{best}{$\blacksquare$}\,best, \textcolor{good}{$\blacksquare$}\,strong, \textcolor{mid!80!black}{$\blacksquare$}\,mid, \textcolor{weak}{$\blacksquare$}\,weak, \textcolor{worst}{$\blacksquare$}\,worst. \sysname wins every column; SlideTalker matches Figure-only on Internal Faithfulness but drops to or below it on Concept Coverage and Takeaway Recall---slide-style narration is correct but doesn't walk through the figure $F$.}
\label{tab:d1-narration}
\end{table*}

\begin{table*}[t]
\centering
\setlength{\tabcolsep}{2.5pt}
\renewcommand{\arraystretch}{1.18}
\begin{tabular}{@{}l cccc@{}}
\toprule
Method &
Inputs &
Mech. &
Outputs &
Takeaway \\
&
\scriptsize CF/CC/EH$\downarrow$ &
\scriptsize CF/CC/EH$\downarrow$ &
\scriptsize CF/CC/EH$\downarrow$ &
\scriptsize CF/CC/EH$\downarrow$ \\
\midrule
\textsc{\sysname}
& \bestcell{.84/.79/.09}
& \bestcell{.79/.74/.12}
& \bestcell{.81/.76/.10}
& \bestcell{.77/.72/.14} \\
SAM+Box
& \highcell{.71/.74/.28}
& \highcell{.59/.61/.34}
& \highcell{.69/.71/.29}
& \highcell{.55/.57/.36} \\
VLM-Grnd.
& \midcell{.58/.62/.34}
& \midcell{.46/.51/.39}
& \midcell{.55/.59/.35}
& \midcell{.42/.46/.41} \\
SlideTalker
& \lowcell{.42/.45/.48}
& \lowcell{.33/.36/.53}
& \lowcell{.40/.43/.50}
& \lowcell{.30/.33/.55} \\
TEA
& \weakcell{.24/.27/.58}
& \weakcell{.17/.20/.66}
& \weakcell{.22/.25/.60}
& \weakcell{.14/.17/.69} \\
C2V-Image
& \weakcell{.21/.24/.61}
& \weakcell{.15/.18/.69}
& \weakcell{.19/.22/.63}
& \weakcell{.12/.15/.72} \\
Veo-3.1
& \weakcell{.18/.21/.62}
& \weakcell{.11/.14/.71}
& \weakcell{.16/.19/.65}
& \weakcell{.09/.12/.74} \\
CogVideoX
& \weakcell{.13/.16/.68}
& \weakcell{.08/.11/.76}
& \weakcell{.12/.15/.70}
& \weakcell{.06/.09/.79} \\
\bottomrule
\end{tabular}
\caption{Grounding Evaluation on the FigTalk-Extended.
Each cell reports \textbf{CF}/\textbf{CC}/\textbf{EH} ($\downarrow$).
Darker shades indicate stronger grounding quality.}
\label{tab:d2b-rubric}
\end{table*}

\begin{table}[t]
\centering
\small
\begin{tabular}{lccc}
\toprule
Configuration & Macro-P & Macro-R & Macro-F1 \\
\midrule
MINARD (full) & \textbf{0.702} & \textbf{0.591} & \textbf{0.642} \\

\midrule
\multicolumn{4}{c}{\textit{Remove grounding constraints}} \\

\hspace{2mm} w/o text grounding & 0.306 & 0.366 & 0.412 \\
\hspace{2mm} w/o visual grounding & 0.452 & 0.541 & 0.592 \\

\hspace{2mm} w/o clustering & 0.641 & 0.578 & 0.608 \\

\bottomrule
\end{tabular}
\caption{Grounding-side ablations (IoU) on FigTalk-Gold. Removing text, visual grounding and clustering weakens temporal alignment.}
\label{tab:grounding_ablation}
\end{table}

\section{Results and Observations}
\label{sec:results}



\begin{figure}[t]
  \centering
  \includegraphics[width=0.48\textwidth]{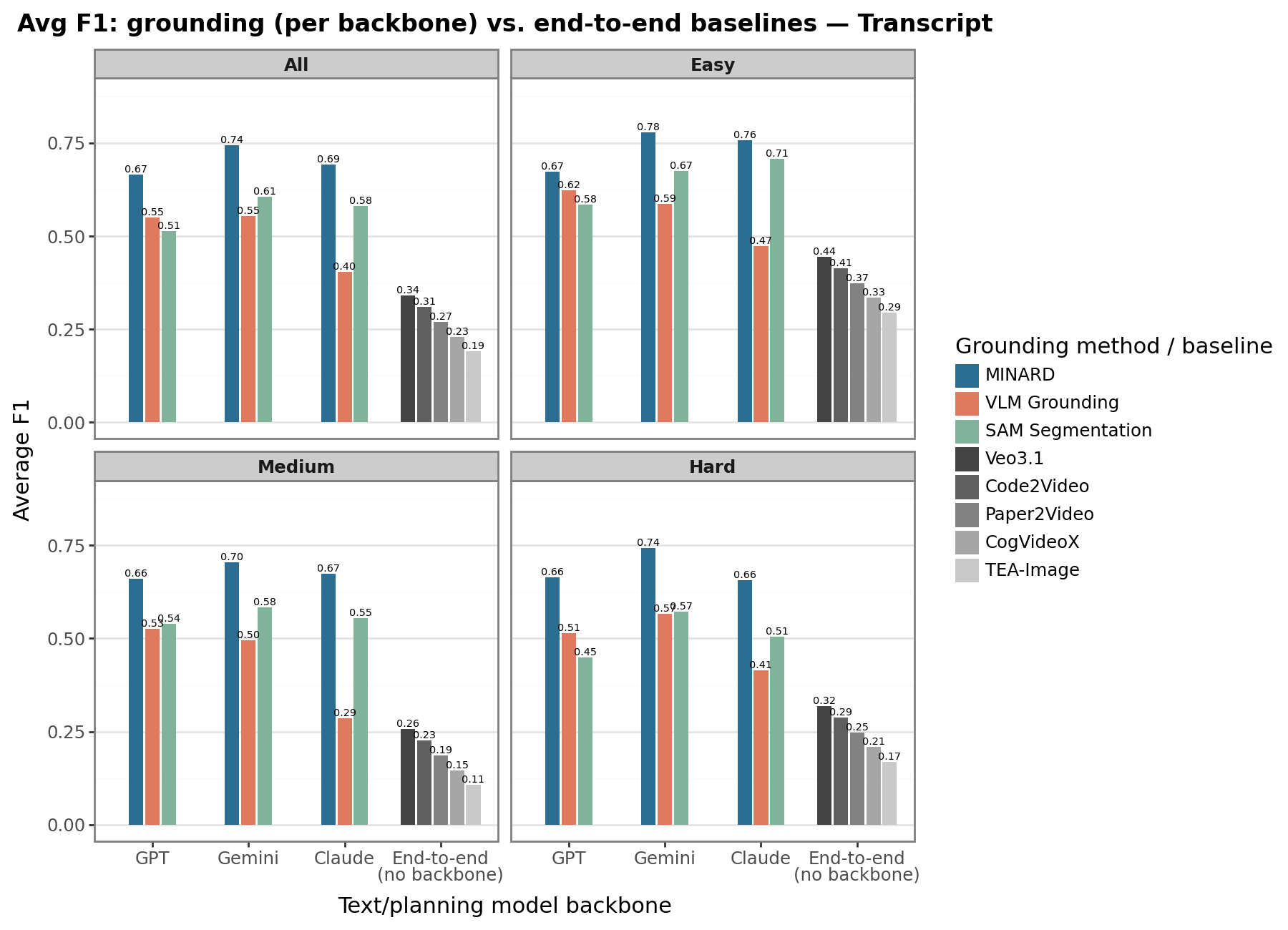}
  \caption{Grounding evaluation on FigTalk-Gold with human narration fixed, stratified by figure complexity levels. We report the average Soft-DTW scores of all baselines against \sysname, showing that \sysname{} achieves substantially stronger performance on the hard-complexity subset compared to existing methods.}
  \label{fig:bar_plot}
\end{figure}

We answer each research question in turn, drawing on the observations based on narration generation (Table~\ref{tab:d1-narration}), grounding (Table~\ref{tab:iou-dtw-stratified}, Figure~\ref{fig:bar_plot} and Table~\ref{tab:d2b-rubric}) and the ablations of \sysname along both the narration quality and grounding quality of \sysname (Table~\ref{tab:narration_ablation2} and Table~\ref{tab:grounding_ablation}).

\paragraph{RQ1: Paper grounding substantially improves narration quality.}
The qualitative examples (Figure~\ref{fig:coin_case_study}) explain the trends in Table~\ref{tab:d1-narration}. Figure-only narration remains strong on \textbf{Order Matching} and \textbf{Internal Faithfulness} because it correctly follows the visual structure and describes visible figure components, but it stays largely descriptive: it never identifies the method name (\textsc{CoIN}), explain the purpose of the contrastive setup, or clarify why the highlighted components matter scientifically. In contrast, \sysname narration recovers missing scientific intent from the paper context, explaining motivation, role of hard negatives, and takeaway of method faithfully, which improves \textbf{Concept Coverage}, \textbf{Takeaway Recall}, and \textbf{External Faithfulness} while preserving strong visual grounding.
The backbone ordering is consistent throughout (\textsc{GPT-5}~$>$~\textsc{Claude}~$>$~\textsc{Gemini}).


\paragraph{RQ2: Structured decomposition and staged grounding win on both
DTW protocols.} We fix the narration to the gold transcript $\{s^*_i\}$
recovered from the human conference video and vary \emph{only} the grounding
mechanism, so every gap reflects grounding alone. We report two complementary metrics on the reference set: \emph{soft-DTW}, a judge-scored frame-level score that places all systems---including the region-free video generators---on one axis (Fig.~\ref{fig:bar_plot}), and \emph{IoU-based DTW}, a stricter region-level score defined only for systems that emit regions (Table~\ref{tab:iou-dtw-stratified}). On both, \textsc{Minard} attains the best score across all backbones; the only exception is \emph{Easy}/GPT, where unconstrained VLM-Grounding edges it on simple single-panel figures. Crucially the gap \emph{inverts and widens} on \emph{Hard} figures
(Fig.~\ref{fig:bar_plot}): \textsc{Minard} stays high while VLM Grounding collapses and the diffusion/animation generators sit near the floor in every tier. On the extended set (Table~\ref{tab:d2b-rubric}), \textsc{Minard} posts the highest Component Faithfulness and Concept Coverage and the \emph{lowest} Excess Highlight Rate on all four axes, with its largest gain on \emph{Mechanism}---baselines over-highlight when uncertain, whereas \textsc{Minard} stays selective. Full metric definitions, per-cell numbers, and the complexity breakdown are
in App.~\ref{app:dtw_details}.

\paragraph{RQ3: The three-team decomposition is critical.} Table~\ref{tab:narration_ablation2} shows that retrieval of paper evidence $K_F$ is the primary driver of scientific faithfulness and takeaway quality: removing retrieval causes the largest collapse in Internal Faithfulness and Takeaway Recall because the narrator falls back to visually descriptive but scientifically shallow explanations. In contrast, the planner and critic-reviser mainly improve pedagogical structure and coherence.
Table~\ref{tab:grounding_ablation} shows that the grounding gains arise from progressively constraining the grounding space. Removing the visual and text grounding from the Perception Team $\Phi$ causes the largest drop because grounding is no longer restricted to valid figure regions. Removing text grounding substantially increases over-highlighting by forcing the selector to search globally over narration steps. Together, these results show that \sysname works because narration, perception, and grounding impose complementary constraints rather than operating as a single unconstrained VLM prediction step.

\section{Human Preference Study}
\label{sec:human-systems}
The metrics in \S\ref{sec:eval-grounding} measure grounding accuracy and narration quality, but not viewer preference. We therefore run two controlled within-group ranking studies with PhD students that isolate the two components of our pipeline: narration generation (Study~I) and grounding quality (Study~II). In each study, one component is fixed and annotators evaluate only the varying axis, preventing confounds. Both studies use 30 figures from 25 papers, uniformly sampled by visual complexity, with five PhD-student rankings per figure. Videos are length-normalized and use the same TTS voice.

\paragraph{Study I: Narration generation (grounding fixed = \textsc{MINARD}).}
With grounding fixed to \textsc{MINARD}, we compare \textsc{N-Summary}, \textsc{N-Step (Figure-only)}, and \textsc{N-Step (Paper+Figure)} on faithfulness, comprehensibility, coherence, and overall preference. \textsc{N-Step (Paper+Figure)} performs best across all criteria (mean rank $1.3$, top-1 rate $74\%$), especially on faithfulness, showing that paper context provides claims unavailable from the figure alone (Kendall's $W=0.66$).

\paragraph{Study II: Grounding model (narration fixed = \textsc{N-Step (Paper+Figure)}).}
Fixing narration to the best Study~I source, we compare \textsc{MINARD}, \textsc{VLM-Grounding}, \textsc{SAM-Grounding}, and \textsc{Code2Video} on excess highlighting, redundant highlighting, step-by-step grounding, and overall preference. \textsc{MINARD} achieves the best grounding quality and overall rank ($1.5$; $63\%$ top-1 rate), while \textsc{Code2Video} ranks last due to frequent over-highlighting of unrelated regions (Kendall's $W=0.72$).

\begin{table}[t]
\centering
\tiny
\caption{\textbf{Study I --- narration generation}, grounding fixed to
\textsc{MINARD}. Only narration-quality criteria are evaluated. Mean rank
($\downarrow$, $1=$ best of three); overall top-1 (\%, $\uparrow$). Best in \textbf{bold}.}
\label{tab:study1}
\setlength{\tabcolsep}{6pt}
\begin{tabular}{lccccc}
\toprule
& \multicolumn{3}{c}{Narration quality} & \multicolumn{2}{c}{Overall} \\
\cmidrule(lr){2-4}\cmidrule(lr){5-6}
Narration source & Faith. & Compr. & Coher. & Rank & Top-1 \\
\midrule
\textsc{N-Summary}            & 2.8 & 2.7 & 2.8 & 2.6 & 9 \\
\textsc{N-Step (Figure-only)} & 2.9 & 2.4 & 2.3 & 2.1 & 17 \\
\textsc{N-Step (Paper+Figure)}& \textbf{1.2} & \textbf{1.4} & \textbf{1.5} & \textbf{1.3} & \textbf{74} \\
\bottomrule
\end{tabular}
\end{table}

\begin{table}[h]
\centering
\tiny
\caption{\textbf{Study II --- grounding model}, narration fixed to
\textsc{N-Step (Paper+Figure)}. Only grounding-quality criteria are evaluated.
Mean rank ($\downarrow$, $1=$ best of four); overall top-1 (\%, $\uparrow$). Best in \textbf{bold}.}
\label{tab:study2}
\setlength{\tabcolsep}{6pt}
\begin{tabular}{lccccc}
\toprule
& \multicolumn{3}{c}{Grounding quality} & \multicolumn{2}{c}{Overall} \\
\cmidrule(lr){2-4}\cmidrule(lr){5-6}
Grounding model & Excess & Redund. & Step & Rank & Top-1 \\
\midrule
\textsc{MINARD}       & \textbf{1.6} & \textbf{1.5} & \textbf{1.3} & \textbf{1.5} & \textbf{63} \\
\textsc{VLM-Grounding}       & 2.3 & 2.4 & 2.6 & 2.3 & 21 \\
\textsc{SAM-Grounding} & 2.9 & 2.9 & 2.8 & 2.8 & 11 \\
\textsc{Code2Video}          & 3.2 & 3.2 & 3.3 & 3.4 & 5  \\
\bottomrule
\end{tabular}
\end{table}

\section{Related Work}
Figure understanding has largely been framed as static interpretation: figure captioning describes what an image contains rather than why a system works\citep{hsu-etal-2021-scicap-generating}, while chart and figure QA target isolated lookups instead of coherent exposition \citep{masry-etal-2022-chartqa}. A complementary line parses diagrams into explicit structure—diagram parse graphs, and diagram-to-text generation beyond chartqa—but recovers structure or answers single queries rather than producing a sequential, narration-conditioned walkthrough. 
Presentation-generation systems such as PPTAgent \citep{zheng-etal-2025-pptagent} and Paper2Video \citep{zhu2025paper2videoautomaticvideogeneration} narrate from papers but treat figures as opaque inserts. 
Unlike programmatic animation systems such as \textsc{TheoremExplainAgent} and Code2Video, or diffusion-based video generators such as Veo and CogVideoX that regenerate content rather than preserve figure fidelity, \sysname{} conditions narration on source paper and grounds each explanation step to explicit, validated regions of original figure. A detailed discussion of related work is in Section~\ref{sec:related-work}.

\section{Conclusion}
We introduce \textsc{FigTalk}, the first benchmark for generating paper-grounded narrations with temporally aligned visual highlights. Our system, \sysname{}, decomposes the task into narration, perception, and grounding contracts, yielding more faithful narrations and human-like grounding traces than existing baselines, with the largest gains on the Hard subset where structured decomposition matters most. Beyond improving accuracy, this modular design also makes failures interpretable. Future work includes extending \sysname{} to interactive and adaptive educational video generation.

\section*{Limitations}
\begin{enumerate}[leftmargin=*]
    \item \sysname{} focuses on architecture-style figures (pipelines, neural architectures, and system overviews) where meaning is conveyed through modules, arrows, and ordered dataflow; charts, plots, and statistical visualizations require different decomposition and narration objectives and are outside the current scope.
    
    \item The pipeline is more computationally expensive than single-pass baselines, with most overhead arising from the critic--reviser refinement loop.
    
    \item The reference grounding set is relatively small (49 videos with gold grounding traces) and derived from existing paper-to-video sources, so results should be interpreted as relative comparisons under a fixed evaluation protocol rather than population-level estimates.
    
    \item Because the soft-DTW protocol relies on a VLM-based judge, its reliability ultimately depends on the quality and calibration of the accompanying human validation.
\end{enumerate}

\section*{Ethical Considerations}
All human evaluation in this work was carried out by student volunteers. 
The annotators were graduate students who participated on a voluntary basis, without monetary or other compensation, and were free to stop or withdraw at any point without consequence. 
No personally identifying information was collected, and annotators reviewed only publicly available scientific figures, papers, and the corresponding generated videos; the task involved no sensitive, private, or otherwise harmful content. We did not store any personally identifiable information from participants, and all students were informed about the research purpose and intended use of the study. The study involved only voluntary participants from within the organization (no crowdsourcing platforms), with no external recruitment or sensitive data collection; therefore, IRB approval was not required.

The source figures and papers used to build FigTalk are drawn from publicly available scientific publications and conference presentation recordings, and we use them solely for the purpose of evaluating figure-explanation quality. 
Because the study involved no intervention beyond reading public scientific material and providing quality judgments, and collected no personal data, it posed minimal risk to participants. We note that our evaluation relies in part on VLM/LLM judges, which may carry biases from their training data; we mitigate this by calibrating against human annotators and reporting agreement, but residual model bias cannot be fully excluded, and automatic scores should be interpreted as approximations of human judgment rather than ground truth. Finally, the system is intended to assist with understanding scientific figures and is not designed for, and should not be relied upon for, autonomous generation of factual scientific claims without human verification.

We used Generative AI (GenAI) in this project. We used Cursor\footnote{https://cursor.com/agents} to make plots, and ChatGPT to refine paper writing, proofreading, and occasional grammatical mistake clarification in framing. GenAI did not directly write any parts of this paper. We take responsibility for GenAI errors. By discussing AI usage here, we encourage the other NLP researchers to do the same.

%

\bibliography{custom}




\section{Appendix Overview}
\label{app:overview}
This appendix is organized as follows.

\begin{figure*}[t]
  \centering
  \includegraphics[width=0.98\textwidth]{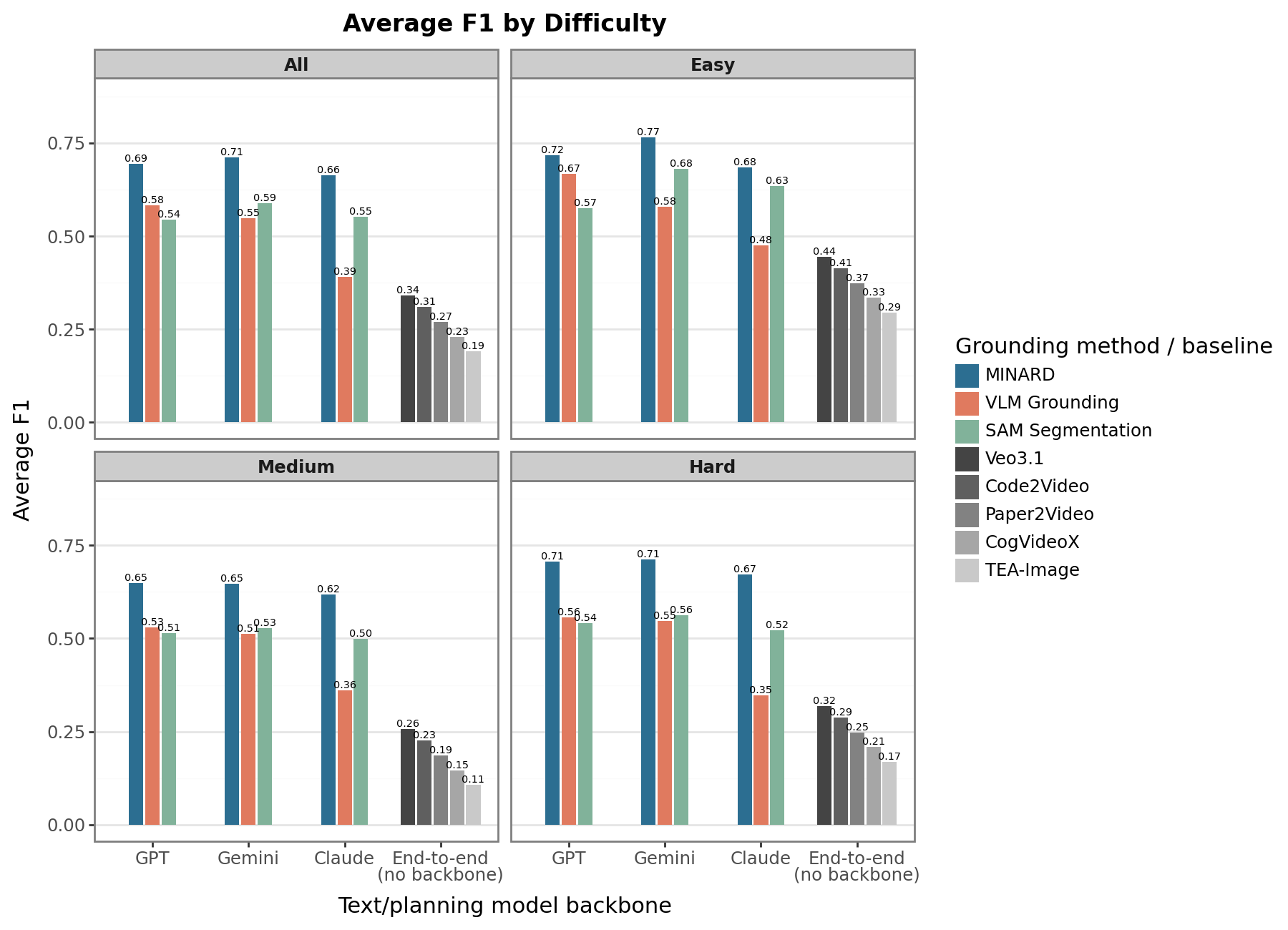}
  \caption{Grounding Evaluation (Using Soft-DTW) on the FigTalk-Gold stratified by the complexity levels showing the Average F1 of various Baselines compared to MINARD, showing that on the Hard-Set MINARD performs significantly better than the existing methods. Here the narration is generated using the figure only and keeping it same we compute the grounding performance.}
  \label{fig:bar_plot_stepbystep}
\end{figure*}

\begin{enumerate}[leftmargin=2.2em]
  \item App.~\ref{app:narration-baselines}: \nameref{app:narration-baselines} presents the Narration Generation Methods.
  \item App.~\ref{app:narration-metrics}: \nameref{app:narration-metrics} presents the Narration Evaluation Measures and Human-VLM Calibration of the Metrics.
  \item App.~\ref{app:grounding-baselines}: \nameref{app:grounding-baselines}
    (incl. App.~\ref{app:baseline-implementations},
    \nameref{app:baseline-implementations}).
  \item App.~\ref{app:grounding-metrics}: \nameref{app:grounding-metrics}.
    \begin{enumerate}
      \item Gold-reference set, scored with
        App.~\ref{app:soft-dtw} (\nameref{app:soft-dtw}) and
        App.~\ref{app:iou-dtw} (\nameref{app:iou-dtw}).
      \item App.~\ref{app:non-gold}: \nameref{app:non-gold}.
    \end{enumerate}
  \item App.~\ref{app:dtw_details}: \nameref{app:dtw_details}.
  \item App.~\ref{app:complexity}: \nameref{app:complexity}.
  \item App.~\ref{app:narr_qual}: \nameref{app:narr_qual}.
  \item App.~\ref{app:prompts}: \nameref{app:prompts}
    (incl. App.~\ref{app:critics}, \nameref{app:critics}).
  \item App.~\ref{app:grounding-impl}: \nameref{app:grounding-impl}.
  \item App.~\ref{sec:narration-ablation}: \nameref{sec:narration-ablation}
    (incl. App.~\ref{sec:discussion} and App.~\ref{sec:cost-analysis}).
  \item App.~\ref{sec:related-work}: \nameref{sec:related-work}.
  \item App.~\ref{app:scope}: \nameref{app:scope}.
  \item App.~\ref{app:prompts}: \sysname{} Implementation Details,
    covering the Retrieval Agent (App.~\ref{app:retrieval}),
    Detection and Segmentation (App.~\ref{app:detect}),
    Clustering and Region Building (App.~\ref{app:builder}),
    and Grounding, Validation, and Rendering (App.~\ref{app:grounding}).
  \item App.~\ref{app:human-eval}: \nameref{app:human-eval}.
  \item Figure~\ref{fig:bar_plot_ishani} and Figure~\ref{fig:bar_plot_stepbystep} shows the grounding performance on FigTalk-Gold when the narrations vary.)  \item Human Evaluation Instructions (App~\ref{app:human_instructions}) 
\end{enumerate}

\section{Human Evaluation Instructions}
\label{app:human_instructions}
We conducted a preference-ranking study to complement the automatic metrics with human judgment of explanation quality. Annotators were shown a scientific figure together with its source paper context (caption and the relevant method paragraphs) and a set of narrated walkthrough videos generated by different systems for that figure, presented in randomized order with system identities hidden. For each figure, annotators were asked to rank the videos from best to worst according to how well each one explains the figure, not how visually polished it is. The instructions defined a good explanation along three criteria, in priority order: (1) grounding accuracy—whether the highlighted regions at each moment correspond to the component being described; (2) explanatory completeness—whether the walkthrough covers the salient modules, their interactions, and the figure's takeaway rather than only naming parts; and (3) ordering coherence—whether the explanation follows a sensible reading order (inputs → modules → outputs → takeaway) without confusing jumps. Annotators were explicitly told to disregard rendering quality, voice, color, and stylistic fluency except where these interfere with understanding, and to treat scientifically equivalent paraphrases as equally valid. Ties were permitted when two videos were genuinely indistinguishable on the criteria above. Each figure was ranked independently by five annotators, and we report inter-rater agreement using Kendall's W over the resulting rankings (Table~\ref{tab:agreement-ranking}). Annotators were graduate students with experience reading, reviewing, or presenting NLP and machine-learning papers, so that they could assess scientific faithfulness rather than only surface fluency. Before the main study, annotators completed a short calibration round on held-out figures and discussed any disagreement on the criteria to align their interpretation; calibration items were excluded from the reported results.
~

\begin{figure}[t]
  \centering
  \includegraphics[width=0.50\textwidth]{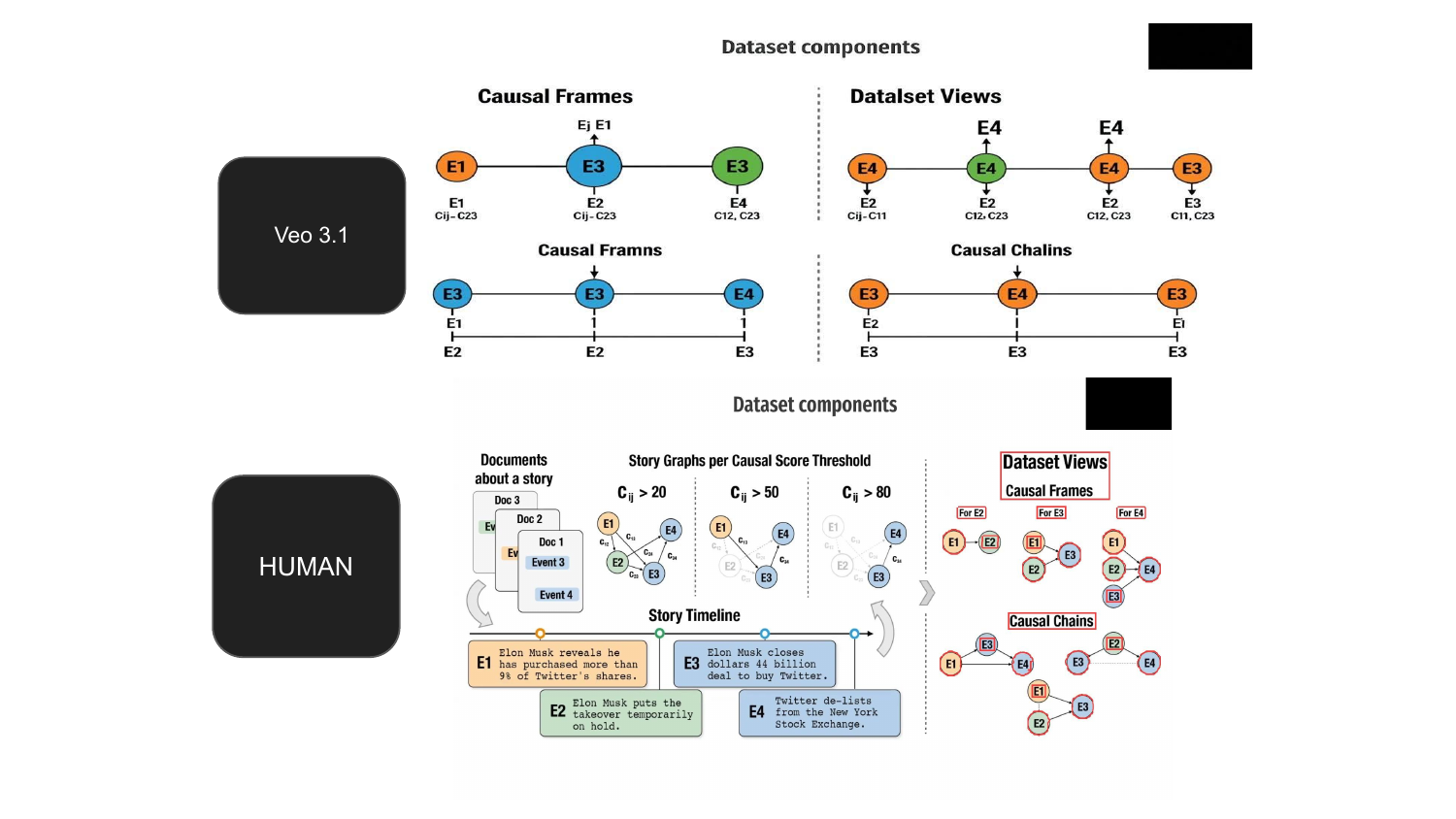}
  \caption{On a structurally complex figure with long-range causal dependencies, Veo~3.1 produces hallucinated figure compared to Gold.}
  \label{fig:example_input_hallu}
  \vspace{-2.25mm}
\end{figure}

\begin{figure*}[t]
  \centering
  \includegraphics[width=0.98\textwidth]{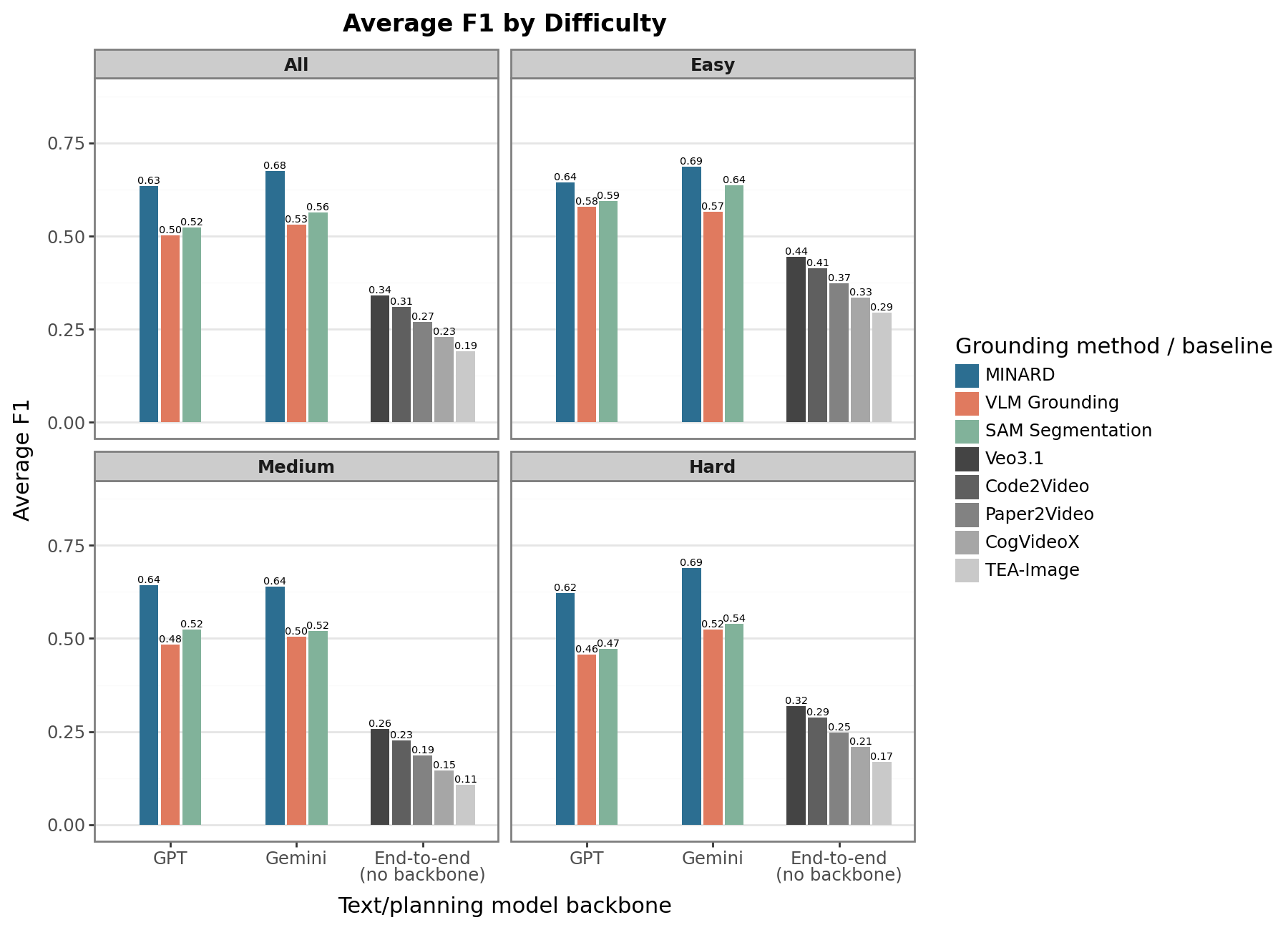}
  \caption{Grounding Evaluation on the FigTalk-Gold (Using Soft-DTW) stratified by the complexity levels showing the Average F1 of various Baselines compared to MINARD, showing that on the Hard-Set MINARD performs significantly better than the existing methods. Here the narration is generated using MINARD-Narration Pipeline from the paper and figure combination  and keeping it same we compute the grounding performance.}
  \label{fig:bar_plot_ishani}
\end{figure*}

\section{Configuration of Narration Baselines}
\label{app:narration-baselines}

\paragraph{Conditioning contexts.} We hold the narration pipeline fixed and vary only the evidence the narrator conditions on, so that within-backbone differences isolate the effect of paper context. \textbf{Figure-grounded ($F$ only)} conditions on the target figure $F$ alone, with no access to the paper, testing whether $F$ is self-sufficient as an explanatory artifact. \textbf{Paper-and-figure-grounded ($F+D$)} conditions on $F$ together with paper context $D$ (introduction, method, caption, and all in-text references to $F$), mirroring a human presenter's access and serving as the default \sysname{} configuration. \textbf{Paper2Video \textsc{Talker-Builder}}~\cite{zhu2025paper2videoautomaticvideogeneration} runs the SlideTalker module, which produces spoken explanations for slide-style content from a paper and an embedded figure; because it treats the figure as an opaque insert and narrates \emph{around} it rather than \emph{through} its parts, its output is closer to a summary than a step-by-step walkthrough, providing a meaningful contrast to \sysname{}. We use the released \textsc{Talker-Builder} prompt unchanged.

\paragraph{Backbones.} Each variant is instantiated with \textsc{Gemini-3.1-Pro}, \textsc{GPT-5}, and \textsc{Claude Sonnet}; within-backbone deltas isolate the contribution of paper context, while within-variant deltas isolate backbone capability.

\paragraph{Prompt for the Figure-only variant.} The Figure-only and Paper-and-Figure variants share the same instruction template and differ only in the evidence block: the Figure-only variant receives only the rendered figure and caption, while the Paper-and-Figure variant additionally includes the retrieved evidence bundle $K_F$ (App.~\ref{app:retrieval}).

\begin{center}
\colorbox{gray!12}{\begin{minipage}{0.95\columnwidth}\small\ttfamily
\textbf{Figure-only Narration Prompt.} SYSTEM: You are an expert author narrating ONE scientific figure to a technical audience, the way a presenter walks through an architecture diagram in a conference talk. TASK: produce an ordered, step-by-step spoken narration that traces the figure as a walkthrough that unfolds over time. READING ORDER: infer the figure structure and data flow, follow a coherent traversal (inputs -> intermediate modules -> outputs -> takeaway), and avoid jumping backward unless necessary. GROUNDING AND FAITHFULNESS: every claim must be supported by the figure or caption; do NOT invent components, datasets, metrics, or training details; preserve the exact figure terminology; explicitly describe module interactions and data flow. COVERAGE: cover all salient modules and major connections, merge redundant elements into coherent steps, and keep each step concise and presentation-like. OUTPUT FORMAT: return ONLY a JSON array [\{"step": 1, "focus": "...", "narration": "..."\}, ...]. INPUT: Figure caption: \{caption\}; [figure image attached].
\end{minipage}}
\end{center}

\section{Narration Evaluation Strategy}
\label{app:narration-metrics}

\paragraph{Overview.}
We evaluate generated narrations using a claim-centric, evidence-grounded framework designed to measure factual correctness, pedagogical structure, explanatory completeness, and scientific usefulness. Unlike lexical-overlap metrics such as BLEU or ROUGE, our evaluation asks whether a narration explains a scientific figure in a logically ordered, faithful, and educationally meaningful way. Given a paper PDF $D$, a target figure $F$, a gold human narration $Y^\star$, and a generated narration $\hat{Y}$, the evaluator computes five complementary dimensions: \textbf{Order Matching}, \textbf{Internal Faithfulness}, \textbf{Concept Coverage}, \textbf{Takeaway Recall}, and \textbf{External Faithfulness}.

The paper and figure are treated as the only sources of factual truth. The gold narration $Y^\star$ is used only for pedagogical metrics that require a reference explanation structure, specifically Order Matching, Concept Coverage, and Takeaway Recall, and is never treated as evidence for factual correctness. Both Internal Faithfulness and External Faithfulness are evaluated directly against evidence from the paper, figure, and external references. This prevents penalizing scientifically correct explanations simply because they differ from one presenter’s walkthrough.

\paragraph{Evidence Construction.}
The evaluation pipeline first constructs a figure-specific evidence bundle $\mathcal{K}_F$ from the source paper. Given a paper $D$ and target figure $F$, we first locate the figure in the PDF and extract its caption using the document layout and figure numbering structure. We then scan the full paper text for explicit references to the figure, including references such as ``Figure 2,'' ``Fig. 2,'' and subfigure mentions like ``Fig. 2(a),'' and collect the surrounding paragraphs as figure-linked explanatory context.

Next, we run OCR over the cropped figure image to recover visible labels, module names, arrows, legends, axes, and other embedded textual elements. We additionally extract nearby explanatory paragraphs around the figure location, since authors often explain the figure immediately before or after it appears. Finally, we collect contribution statements from the abstract, introduction, and conclusion, together with the local section and subsection context containing the figure. The resulting bundle $\mathcal{K}_F$ therefore combines caption text, figure mentions, OCR content, nearby explanatory paragraphs, contribution statements, and section-level context, and serves as the grounding evidence for all paper- and figure-grounded factual verification.

\paragraph{Narration Decomposition.}
Both the gold narration and generated narration are segmented into ordered narration steps. Each step is then decomposed into atomic factual claims using a structured VLM/LLM-based extraction procedure. Claims correspond to independently verifiable scientific statements such as describing a module's role, architectural dependency, information flow, experimental finding, causal relationship, or scientific takeaway.

The claim extraction stage instructs the evaluator to identify all independently verifiable scientific statements while avoiding rhetorical or stylistic language. Compound statements are separated into multiple atomic claims whenever different parts can be independently verified. The extracted representation contains the claim text, referenced concepts, claim type, and supporting narration span. Claim types include method descriptions, architectural behavior, results, definitions, background knowledge, takeaway statements, and causal relationships.

\begin{center}
\colorbox{gray!12}{\begin{minipage}{0.96\columnwidth}\small\ttfamily
You are given one narration step from a scientific figure explanation video together with the associated figure and paper context. Extract all atomic factual claims expressed in the narration step.

An atomic factual claim should correspond to exactly one independently verifiable scientific statement. Separate compound statements whenever different parts could be independently verified. Preserve the scientific meaning of the narration while avoiding rhetorical language or stylistic phrasing.

For each extracted claim, identify:
(1) the exact claim text,
(2) referenced scientific concepts,
(3) claim type,
(4) supporting narration span.

Claim types include:
\{method, architecture, result, definition, background, takeaway, causal\_relation\}.

Do not infer unstated scientific details. Extract only claims explicitly supported by the narration.

Return structured JSON only.
\end{minipage}}
\end{center}

\paragraph{Internal Faithfulness.}
Internal Faithfulness evaluates whether generated narration claims are supported by the paper and figure evidence bundle $\mathcal{K}_F$. The evaluator verifies each extracted claim against figure captions, OCR text, visual modules, nearby paragraphs, figure references, and contribution statements. Importantly, the gold narration is never used as factual evidence, since it represents only one valid explanatory walkthrough rather than an oracle for truth.

The evaluation prompt explicitly instructs the judge to treat semantic paraphrases as faithful whenever they preserve the same scientific meaning, while penalizing fabricated modules, unsupported causal reasoning, incorrect architectural behavior, hallucinated relationships, exaggerated claims, and incorrect result interpretations. The evaluator additionally produces evidence spans, rationales, and confidence scores for every prediction.

\begin{center}
\colorbox{gray!12}{\begin{minipage}{0.96\columnwidth}\small\ttfamily
You are evaluating whether a generated scientific narration claim is factually supported by the paper and figure evidence.

You are given:
(1) a generated narration claim,
(2) figure evidence including OCR text, captions, visual modules, and figure references,
(3) relevant paper paragraphs and contribution statements.

Determine whether the claim is:
\{entailed, contradicted, not\_supported\}.

Use only the supplied paper and figure evidence. Do not use outside knowledge. Do not use the gold narration as evidence.

A claim is entailed only if the supplied evidence directly supports the scientific meaning of the claim.

Penalize fabricated modules, incorrect architectural behavior, unsupported causal reasoning, exaggerated claims, incorrect result interpretations, and hallucinated relationships.

Return:
(1) label,
(2) supporting evidence spans,
(3) rationale,
(4) confidence score.

Return structured JSON only.
\end{minipage}}
\end{center}

\paragraph{Order Matching.}
Order Matching evaluates whether the narration follows a pedagogically coherent explanation sequence. To support this evaluation, we construct a gold explanation graph from the human narration and paper context, where nodes correspond to concepts and edges encode prerequisite or explanatory dependencies.

The evaluator identifies the primary concept discussed in each narration step and determines whether the transition from the previous step represents a forward explanation, an acceptable recap, or a dependency violation. The instructions explicitly note that scientific presentations frequently revisit previously introduced concepts for clarification, and therefore recaps should not be penalized as strongly as genuine prerequisite violations.

\begin{center}
\colorbox{gray!12}{\begin{minipage}{0.96\columnwidth}\small\ttfamily
You are evaluating whether a scientific narration follows a pedagogically coherent explanation order.

You are given:
(1) the current narration step,
(2) the previous narration step,
(3) a gold explanation graph describing prerequisite relationships between concepts.

Identify the primary scientific concept explained in the current step and determine whether the transition from the previous step represents:
\{forward, recap, violation\}.

A recap revisits a previously introduced concept for clarification or reinforcement and should receive partial credit.

Do not over-penalize minor local reorderings if the overall pedagogical flow remains coherent.

Return:
(1) identified concept,
(2) transition label,
(3) rationale,
(4) confidence score.

Return structured JSON only.
\end{minipage}}
\end{center}

\paragraph{Concept Coverage.}
Concept Coverage evaluates whether the narration meaningfully explains the important scientific concepts associated with the figure and paper. The target concept set is constructed from the gold narration, figure caption, contribution statements, and figure-linked paper context.

The evaluator explicitly distinguishes between merely naming a module and actually explaining its scientific role. A concept is considered covered only if the narration refers to the concept or a valid semantic alias and additionally explains at least one scientifically correct role, dependency, interaction, or relationship involving the concept.

\begin{center}
\colorbox{gray!12}{\begin{minipage}{0.96\columnwidth}\small\ttfamily
You are evaluating whether a generated narration meaningfully explains an important scientific concept associated with the figure and paper.

A concept is covered only if:
(1) the narration refers to the concept or a valid semantic alias,
and
(2) the narration explains at least one scientifically correct relationship or role involving the concept.

A superficial mention without explanation should not receive coverage credit.

Recognize scientifically valid paraphrases and synonymous terminology.

Distinguish between shallow module naming and meaningful scientific explanation.

Return:
(1) coverage label,
(2) matched alias,
(3) supporting narration span,
(4) explained relationship,
(5) rationale,
(6) confidence score.

Return structured JSON only.
\end{minipage}}
\end{center}

\paragraph{Takeaway Recall.}
Takeaway Recall measures whether the narration preserves the central scientific conclusions and reasoning conveyed by the figure and paper. The evaluator compares the generated narration against takeaway statements extracted from the paper contributions, conclusions, and human narration summaries.

The evaluation instructions emphasize that the judge should determine whether the narration preserves not only the final scientific conclusion but also the supporting reasoning or causal explanation behind the conclusion. Partial credit is assigned when the narration captures the high-level takeaway while omitting important rationale details.

\begin{center}
\colorbox{gray!12}{\begin{minipage}{0.96\columnwidth}\small\ttfamily
You are evaluating whether a generated narration preserves the central scientific takeaway of a figure explanation.

You are given:
(1) gold takeaway statements,
(2) rationale statements explaining why the takeaway follows,
(3) the generated narration.

Determine whether the generated narration preserves:
(a) the primary scientific conclusion,
and
(b) the supporting reasoning or causal explanation.

Classify the result as:
\{entailed, partially\_entailed, not\_entailed\}.

Semantic paraphrases and alternative pedagogically valid explanations should still count as correct if they preserve the same scientific meaning.

Return:
(1) label,
(2) matched generated claims,
(3) missing rationale components,
(4) rationale,
(5) confidence score.

Return structured JSON only.
\end{minipage}}
\end{center}

\paragraph{External Faithfulness.}
External Faithfulness evaluates whether additional background knowledge introduced by the narration is scientifically correct. The evaluator first identifies claims that are not directly supported by the paper or figure but correspond to external scientific context such as textbook definitions, descriptions of standard modules, or commonly accepted domain knowledge.

These claims are verified against an external evidence bundle $\mathcal{K}_{\mathrm{ext}}$ retrieved from cited references and curated scientific resources. The evaluator is explicitly instructed not to penalize scientifically correct clarifications simply because they are absent from the source paper while still rejecting unsupported or misleading background claims.

\begin{center}
\colorbox{gray!12}{\begin{minipage}{0.96\columnwidth}\small\ttfamily
You are evaluating whether externally introduced scientific background information is correct.

You are given:
(1) a generated narration claim not directly supported by the paper,
(2) external reference evidence retrieved from cited papers, textbooks, or curated scientific resources.

Determine whether the claim is:
\{externally\_supported, externally\_contradicted, not\_verifiable\}.

Do not penalize scientifically correct clarifications simply because they are absent from the source paper.

Reject unsupported scientific generalizations, fabricated background knowledge, incorrect definitions, or misleading explanations.

Return:
(1) label,
(2) supporting reference evidence,
(3) rationale,
(4) confidence score.

Return structured JSON only.
\end{minipage}}
\end{center}

\paragraph{VLM/LLM Judge Calibration.}
All evaluation dimensions are implemented using structured VLM/LLM judges operating over multimodal scientific evidence. We evaluate multiple judge models, including \textsc{Gemini-2.5-Flash}, \textsc{Gemini-2.5-Pro}, \textsc{GPT-4o}, \textsc{Claude-3.7-Sonnet}, and \textsc{DeepSeek-R1}. Each judge receives identical prompts, identical evidence bundles, identical figure context, and identical JSON output constraints so that differences reflect judge quality rather than prompt engineering or formatting inconsistencies.

To calibrate the judges, we used the exact same inputs, evidence bundles, prompt instructions, and scoring criteria provided to the VLM/LLM evaluators and asked a PhD student with prior experience reading, reviewing, and presenting NLP and machine learning papers to independently perform the same evaluation procedure. The annotator assigned structured judgments across all five evaluation dimensions: Order Matching, Internal Faithfulness, Concept Coverage, Takeaway Recall, and External Faithfulness.

The annotator received the same narration inputs, figure crops, OCR text, captions, figure-linked paper context, and evaluation instructions used by the automatic judges. The instructions emphasized evaluating scientific faithfulness and pedagogical quality rather than stylistic fluency. The annotator was explicitly instructed not to penalize semantic paraphrases, local reorderings, or scientifically correct clarifications absent from the gold narration, while distinguishing between factual hallucinations, unsupported causal reasoning, shallow module naming, and genuinely meaningful scientific explanation.

The calibration set consisted of papers spanning neural architectures, multimodal systems, training pipelines, reasoning workflows, and experimental diagrams. For each paper, multiple generated narrations produced using different prompting strategies and generation systems were evaluated independently by both the human annotator and the VLM/LLM judges.

We then measured agreement between human judgments and VLM/LLM outputs across all five evaluation dimensions using Kendall's $\tau$. Table~\ref{tab:judge-calibration} reports the resulting human--VLM agreement statistics. Among all evaluated judges, \textsc{Gemini-2.5-Flash} achieved the strongest agreement with the human annotator across all evaluation dimensions while additionally producing the most stable rationales and lowest structured-output failure rate. We therefore use \textsc{Gemini-2.5-Flash} as the primary evaluator throughout all experiments.

\begin{table*}[t]
\centering
\small
\begin{tabular}{lcccccc}
\toprule
\textbf{Judge} & \textbf{Order} & \textbf{Internal} & \textbf{Coverage} & \textbf{Takeaway} & \textbf{External} & \textbf{JSON Fail} \\
 & \textbf{$\tau$} & \textbf{$\tau$} & \textbf{$\tau$} & \textbf{$\tau$} & \textbf{$\tau$} & \\
\midrule
Gemini-2.5-Flash & \textbf{0.72} & \textbf{0.76} & \textbf{0.70} & \textbf{0.68} & \textbf{0.66} & \textbf{0.4\%} \\
Gemini-2.5-Pro & 0.68 & 0.72 & 0.67 & 0.63 & 0.61 & 0.8\% \\
GPT-4o & 0.64 & 0.69 & 0.63 & 0.60 & 0.57 & 1.6\% \\
Claude-3.7-Sonnet & 0.61 & 0.66 & 0.60 & 0.57 & 0.54 & 1.3\% \\
DeepSeek-R1 & 0.55 & 0.60 & 0.56 & 0.52 & 0.49 & 3.8\% \\
\bottomrule
\end{tabular}
\caption{Human--VLM agreement across all narration evaluation dimensions using Kendall's $\tau$. JSON Fail measures malformed structured outputs.}
\label{tab:judge-calibration}
\end{table*}

\section{Grounding Metrics: Full Definitions}
\label{app:grounding-metrics}

\begin{table*}[t]
\centering
\small
\setlength{\tabcolsep}{5pt}
\begin{tabular}{lll ccc}
\toprule
Tier & Backbone & Grounding method & Macro-F1 & Macro-P & Macro-R \\
\midrule
\multicolumn{6}{l}{\textit{All}} \\
\midrule
\multirow{9}{*}{All}
 & \multirow{3}{*}{GPT}    & \sysname          & \textbf{0.666} & 0.723 & 0.712 \\
 &                         & VLM-Grounding   & 0.593 & 0.600 & 0.708 \\
 &                         & SAM Seg + BBox  & 0.521 & 0.732 & 0.487 \\
\cmidrule(l){2-6}
 & \multirow{3}{*}{Gemini} & \sysname          & \textbf{0.743} & 0.783 & 0.787 \\
 &                         & SAM Seg + BBox  & 0.625 & 0.792 & 0.617 \\
 &                         & VLM-Grounding   & 0.554 & 0.620 & 0.579 \\
\cmidrule(l){2-6}
 & \multirow{3}{*}{Claude} & \sysname          & \textbf{0.693} & 0.736 & 0.751 \\
 &                         & SAM Seg + BBox  & 0.670 & 0.797 & 0.668 \\
 &                         & VLM-Grounding   & 0.404 & 0.528 & 0.386 \\
\midrule
\multicolumn{6}{l}{\textit{Easy}} \\
\midrule
\multirow{9}{*}{Easy}
 & \multirow{3}{*}{GPT}    & VLM-Grounding   & \textbf{0.749} & 0.764 & 0.830 \\
 &                         & \sysname          & 0.673 & 0.751 & 0.705 \\
 &                         & SAM Seg + BBox  & 0.584 & 0.808 & 0.543 \\
\cmidrule(l){2-6}
 & \multirow{3}{*}{Gemini} & \sysname          & \textbf{0.778} & 0.819 & 0.808 \\
 &                         & SAM Seg + BBox  & 0.675 & 0.862 & 0.647 \\
 &                         & VLM-Grounding   & 0.587 & 0.680 & 0.599 \\
\cmidrule(l){2-6}
 & \multirow{3}{*}{Claude} & \sysname          & \textbf{0.758} & 0.782 & 0.804 \\
 &                         & SAM Seg + BBox  & 0.741 & 0.839 & 0.736 \\
 &                         & VLM-Grounding   & 0.474 & 0.658 & 0.432 \\
\midrule
\multicolumn{6}{l}{\textit{Medium}} \\
\midrule
\multirow{9}{*}{Medium}
 & \multirow{3}{*}{GPT}    & \sysname          & \textbf{0.660} & 0.685 & 0.727 \\
 &                         & SAM Seg + BBox  & 0.564 & 0.691 & 0.559 \\
 &                         & VLM-Grounding   & 0.525 & 0.507 & 0.665 \\
\cmidrule(l){2-6}
 & \multirow{3}{*}{Gemini} & \sysname          & \textbf{0.704} & 0.725 & 0.769 \\
 &                         & SAM Seg + BBox  & 0.653 & 0.743 & 0.690 \\
 &                         & VLM-Grounding   & 0.496 & 0.550 & 0.522 \\
\cmidrule(l){2-6}
 & \multirow{3}{*}{Claude} & \sysname          & \textbf{0.675} & 0.713 & 0.753 \\
 &                         & SAM Seg + BBox  & 0.644 & 0.740 & 0.673 \\
 &                         & VLM-Grounding   & 0.286 & 0.367 & 0.279 \\
\midrule
\multicolumn{6}{l}{\textit{Hard}} \\
\midrule
\multirow{9}{*}{Hard}
 & \multirow{3}{*}{GPT}    & \sysname          & \textbf{0.665} & 0.726 & 0.708 \\
 &                         & VLM-Grounding   & 0.524 & 0.541 & 0.648 \\
 &                         & SAM Seg + BBox  & 0.448 & 0.703 & 0.402 \\
\cmidrule(l){2-6}
 & \multirow{3}{*}{Gemini} & \sysname          & \textbf{0.742} & 0.789 & 0.782 \\
 &                         & SAM Seg + BBox  & 0.572 & 0.772 & 0.551 \\
 &                         & VLM-Grounding   & 0.566 & 0.620 & 0.601 \\
\cmidrule(l){2-6}
 & \multirow{3}{*}{Claude} & \sysname          & \textbf{0.656} & 0.716 & 0.714 \\
 &                         & SAM Seg + BBox  & 0.632 & 0.795 & 0.618 \\
 &                         & VLM-Grounding   & 0.414 & 0.518 & 0.407 \\
\bottomrule
\end{tabular}
\caption{IoU-based DTW grounding on the reference set, stratified by figure
complexity, with narration fixed to the gold transcript $\{s^*_i\}$. Higher is better; best Macro-F1 per (tier, backbone) cell in bold. \sysname uses its strongest planner/traversal configuration per backbone; baselines use the unified backbone. \sysname wins every cell except Easy/GPT, where unconstrained VLM-Grounding edges it on simple single-panel figures; the gap inverts and widens in \sysname's favor on Hard figures. The cross-family soft-DTW comparison (all systems, including video-generation baselines) is in Figure~\ref{fig:bar_plot}.}
\label{tab:iou-dtw-stratified}
\end{table*}


\begin{table}[t]
\centering
\tiny
\setlength{\tabcolsep}{4pt}
\begin{tabular}{lcc}
\toprule
Sub-metric & Inter-human $\kappa$ & VLM--human $\kappa$ \\
\midrule
\multicolumn{3}{l}{\emph{$D_2(b)$ -- Grounding Rubric}} \\
Inputs    & 0.80 & 0.73 \\
Mechanism & 0.74 & 0.69 \\
Outputs   & 0.82 & 0.75 \\
Takeaway  & 0.83 & 0.77 \\
\bottomrule
\end{tabular}
\caption{Agreement on the \textsc{FigTalk} validation sample, stratified by complexity). Inter-human: five PhD NLP annotators using the rubric. VLM--human: agreement between the VLM judge and the average human label per item.}
\label{tab:agreement-human}
\end{table}

\subsection{Determining Grounding Performance on Gold-Reference Set of FigTalk}
We score a system's grounding against the human presenter by treating both as ordered sequences of steps, where each step carries a set of highlighted figure regions, and measuring how well the system sequence matches the human one. The gold sequence is $G = [G_1, \dots, G_n]$ and the system sequence is $P = [P_1, \dots, P_m]$, where each $G_i$ or $P_j$ is a set of regions and a region is a labeled bounding box (a text label, a visual element, or a cluster). The two sequences need not have the same length, since a system may split one human step into several or merge several into one. We use two metrics, both returning a score in $[0,1]$ where higher is better: \textbf{soft-DTW} (\S\ref{app:soft-dtw}), which works for every system including video-generation systems that output no regions and is therefore the single cross-family number in Figure~\ref{fig:bar_plot}, and \textbf{IoU-based DTW} (\S\ref{app:iou-dtw}), which is more precise but defined only for systems that output explicit regions and is reported per backbone and complexity tier in Table~\ref{tab:iou-dtw-stratified}.

\paragraph{Gold sequence construction.} We obtain $G$ from each human conference video by sampling frames at the moments where the presenter's highlight changes and recording, at each such moment, the set of regions that newly turn on; this set is one gold step $G_i$. Two authors do this labeling independently for every video, marking both where each step begins and which regions belong to it, and resolve any disagreement on step boundaries or region membership by discussion. We keep only the 49 videos that pass the grounding audit of \S2 (Cohen's $\kappa = 0.81$), since the gold sequence is the most sensitive object in the evaluation and weak videos would only add noise.

\subsubsection{Soft-DTW (cross-family score)}
\label{app:soft-dtw}

Soft-DTW compares each gold step to each system frame using a VLM judge, so it applies even to systems that produce frames rather than regions such as Veo, Code2Video, CogVideoX. 
For each gold step $i$ and sampled frame $j$, the judge is shown the gold figure with step $i$'s regions drawn as red boxes alongside frame $j$, and returns an integer similarity from $1$ (no match) to $10$ (perfect match), which we convert to a cost $c(i,j) = (10 - \mathrm{score})/9$ so that a score of $10$ costs $0$ and a score of $1$ costs $1$. We then run dynamic time warping over the full step-by-frame cost matrix using $D(i,j) = c(i,j) + \min(D(i{-}1,j),\, D(i,j{-}1),\, D(i{-}1,j{-}1))$, where the three terms advance the gold step, advance the frame, or advance both; the path starts at the first step and frame, ends at the last step and frame, and is recovered by backtracking. The reported score is $1$ minus the mean cost along the aligned path. Because this needs only frames, we apply it to every system: for systems that output regions we render their highlights onto the figure to make frames, putting all systems on one axis, which is the number plotted in Figure~\ref{fig:bar_plot} and the prompt to VLM is below

\begin{center}
\colorbox{gray!12}{%
\begin{minipage}{0.95\columnwidth}
\tiny\ttfamily
\textbf{Soft-DTW Judge Prompt}\\[4pt]
You are a strict visual highlighted-region match scorer.\\[2pt]
You will receive two images:\\
1. Image 1 is the ground-truth image. Its highlighted or emphasized region is the target.\\
2. Image 2 is the prediction image. Its highlighted or emphasized region is the prediction.\\[2pt]
A highlighted region means any region that is visually emphasized or focused on, including a bounding box, colored overlay, crop, zoom, pointer, mask, isolation, or other clear visual focus. A region being visible does not mean it is highlighted.\\[2pt]
Score how closely the highlighted region in Image 2 matches the highlighted region in Image 1.\\[2pt]
Be very strict:\\
Give high scores only when the prediction highlights the same region as the ground truth.\\
Penalize missing any ground-truth highlighted area.\\
Penalize highlighting the wrong region.\\
Penalize highlighting a region that is much larger than the ground-truth region.\\
Penalize any incorrect or irrelevant region that is also included inside the predicted highlighted/focused area.\\
Apply a strong penalty when the predicted highlighted/focused region contains a lot of incorrect or irrelevant content, even if it also contains the correct target.\\
Do not reward the prediction merely because the correct target is visible somewhere in the image. It must be the highlighted/focused region.\\
Ignore style, color, rendering quality, and topic similarity unless the highlighted/focused regions match.\\
When unsure between two scores, choose the lower score.\\[2pt]
Rubric:\\
10 = the predicted highlighted region matches the ground-truth highlighted region almost exactly, with no meaningful irrelevant content\\
8 = the predicted highlighted region clearly matches the target, with only minor extra surrounding content\\
6 = the predicted highlighted region mostly covers the target, but is loose, incomplete, or includes some irrelevant content\\
4 = the predicted highlighted region overlaps the target but includes substantial irrelevant content or misses substantial target content\\
2 = the prediction highlights the wrong region, or the target is only visible but not actually highlighted/focused\\
1 = unrelated, no visible target support, or no clear predicted highlight/focus\\[2pt]
Return only JSON with:\\
\{"score": <integer 1-10>, "rationale": "<brief reason>"\}
\end{minipage}%
}
\end{center}

\paragraph{Judge calibration.} Because soft-DTW depends on VLM judge, we calibrate it against humans before fixing the final model. On a small stratified sample of videos, A PhD-student annotator and each candidate judge (Gemini-3-Flash, GPT) score the same (gold step, frame) pairs on the $1$--$10$ scale, and we measure agreement between each VLM judge and the human consensus as well as among the humans themselves. The judges show high agreement with humans (Gemini $\kappa = \texttt{0.76}$, GPT $\kappa = \texttt{0.67}$) and we select the judge with the strongest human agreement as the VLM-Judge for all reported soft-DTW results. This keeps the automatic score anchored to human judgment.

\begin{figure*}[t]

\centering

\includegraphics[width=0.98\linewidth]{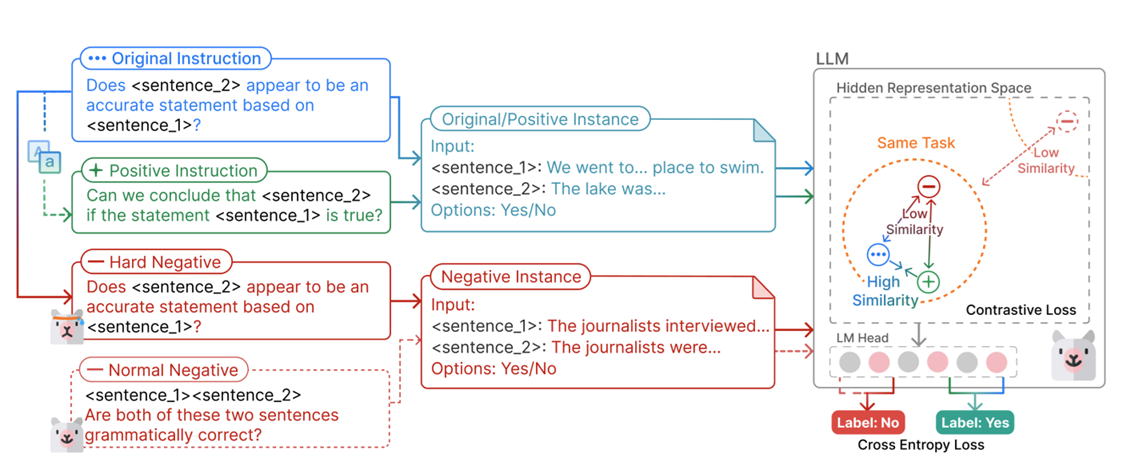}

\vspace{0.4em}



\vspace{0.5em}

\scriptsize
\renewcommand{\arraystretch}{1.15}

\begin{tabular}{@{}p{0.18\linewidth}p{0.77\linewidth}@{}}
\toprule
\textbf{Regime} & \textbf{Generated narration (\textsc{GPT-5})} \\
\midrule

\textbf{Figure-only ($F$)}
&
Let's take a closer look at this figure, which visually captures a method for improving how large language models represent and differentiate between inputs. Starting at the top, the idea is clearly stated: the goal is to encourage semantically equivalent inputs to stay close to each other in the model's hidden representation space, while dissimilar inputs are pushed farther apart.

\smallskip

Moving down, we see four types of instructions labeled: `Original Instruction', `Positive Instruction', `Hard Negative', and `Normal Negative'. Each of these instructions is paired with corresponding instances. For example, the `Original Instruction' and `Positive Instruction' lead to what's called `Original/Positive Instance', while the `Hard Negative' and `Normal Negative' lead to `Negative Instance'. These instances serve as inputs to the LLM, which is shown in the center of the figure.

\smallskip

Now, as we focus on the `Hidden Representation Space', we see that it plays a critical role in separating inputs based on their semantic similarity. The figure highlights `High Similarity' and `Low Similarity' for inputs within the same task, and also introduces the concepts of `Far OODs' and `Near OODs'. `Far OODs' represent examples that are far out-of-distribution, while `Near OODs' are closer but belong to a different class.

\smallskip

Moving further to the right, we see how the `Hidden Representation Space' feeds into `Contrastive Loss', which helps enforce these separations, while the `LM Head' outputs labels that are processed using `Cross Entropy Loss'. The key visual point here is that this method integrates contrastive learning to improve the model's ability to distinguish between inputs based on their semantic relationships.
\\

\midrule

\textbf{Paper+Figure ($F{+}D$)}
&
One of the biggest hurdles for the trustworthiness of LLMs is their sensitivity to phrasing; a simple paraphrase of an instruction can often lead to wildly inconsistent outputs. To address this, we developed \cellcolor{green!12}\textsc{CoIN}, or Contrastive Instruction Tuning.

\smallskip

If you look at Figure~2, you can see the core idea: we want to force the model's hidden representation space to group semantically equivalent inputs together while pushing dissimilar ones apart. On the left side of the slide, we show how we construct our training batches. We take an `Original Instruction' and pair it with a `Positive Instruction'---which is a semantically equivalent paraphrase---using the same input instance.

\smallskip

To provide a contrast, we also introduce \cellcolor{green!12}`Hard Negatives' and `Normal Negatives.' A key point here is the Hard Negative: it uses the exact same instruction as the original but is paired with a different instance from the same task that has a different label.

\smallskip

Moving to the center, these are all processed by the LLM. On the right, you can see the resulting `Hidden Representation Space.' We apply a Contrastive Loss to ensure the original and positive pairs have high similarity, while the negatives---especially those `Near OODs'---are pushed further away.

\smallskip

\cellcolor{green!12}We found that standard negatives from different tasks, or `Far OODs,' are already quite distinct to the model, so focusing on these hard negatives provides the critical supervision signal needed for robustness.

\smallskip

Finally, we combine this contrastive objective with the standard Cross Entropy Loss at the LM head. What I want you to remember is that by explicitly aligning these hidden representations, \textsc{CoIN} allows the model to recognize the underlying intent of an instruction regardless of its textual form, leading to a significant boost in robustness across phrasing variations.
\\

\bottomrule
\end{tabular}

\vspace{0.5em}


\caption{
Comparison between narration generated using only the figure versus narration generated using both the paper and figure. 
While both narrations correctly follow the visual structure, the figure-only setting remains largely descriptive and misses the scientific motivation and intended takeaway. 
Adding paper context recovers the method name (\textsc{CoIN}), explains the hard-negative design choice, and substantially improves Concept Coverage, Takeaway Recall, and External Faithfulness. Both narrations trace the same left-to-right structure faithfully, so Order Matching and Internal Faithfulness remain high in both settings. However, the figure-only narration stays largely descriptive: it lists modules and labels verbatim, never identifies the method name \textsc{CoIN}, and fails to explain why hard negatives matter. As a result, Concept Coverage improves only partially and Takeaway Recall remains weak.
In contrast, the paper-grounded narration injects the missing scientific intent. It explicitly frames the robustness motivation, explains the role of semantically equivalent paraphrases, clarifies why hard negatives provide stronger supervision than far-OOD negatives, and concludes with the intended takeaway about robustness to instruction phrasing.
}
\label{fig:coin_case_study}
\end{figure*}

\subsubsection{IoU-based DTW (Region-level score)}
\label{app:iou-dtw}

For systems that output explicit regions, we compare region sets directly, which is more precise than the frame-based judge. For a gold step $G_i$ and a system step $P_j$, we match each predicted region to a gold region by bounding-box overlap, allowing text regions to match at a lower overlap threshold when their OCR text agrees and matching visual regions by spatial overlap alone. From these matches we count true positives (matched pairs), false positives (predicted regions with no gold match), and false negatives (gold regions with no predicted match), and compute $P = \mathrm{TP}/(\mathrm{TP}+\mathrm{FP})$, $R = \mathrm{TP}/(\mathrm{TP}+\mathrm{FN})$, $F_1 = 2PR/(P+R)$, and a per-step cost $c(G_i,P_j) = 1 - F_1$, so a perfect region match costs $0$ and no overlap costs $1$. We align the sequences with the same recurrence as \S\ref{app:soft-dtw} using this $F_1$ cost, and when several system steps align to one gold step we union their predicted regions before scoring, which credits a system that spreads the correct regions across neighboring steps. We then compute $F_1$ once per gold step and macro-average over gold steps so each step counts equally and a few large steps cannot dominate, reporting Macro-P, Macro-R, and Macro-F1. This metric cannot be computed for video-generation systems, which output no regions.

Both metrics share the same alignment, the same $[0,1]$ scale, the same per-step macro-averaging, and the same direction, differing only in the cost: IoU-based DTW measures exact region overlap and is precise but region-only, while soft-DTW measures judged frame similarity and works for everyone but is coarser. We do not average or swap them, because soft-DTW also depends on the frame-sampling rate and on the judge, which region overlap does not, so a soft-DTW value and an IoU-based value are not directly comparable to each other even when both are high.



\subsection{Determining Grounding Performance of Non-Gold Set of FigTalk}
\label{app:non-gold}
For each of the four axes (Inputs / Mechanism / Outputs / Takeaway), the three judgments are a) \textbf{Component Faithfulness}: for every component highlighted within the axis's scope, the simultaneous narration span is checked against $F$ (location and connectivity) and against $D$ (role attribution). Scored at the (highlight, narration-span) pair level and aggregated to the axis, b) \textbf{Concept Coverage} within the axis's scope, the fraction of gold-narration concepts assigned to that axis that are highlighted-and-explained in the rendered video, c) \textbf{Excess Highlight Rate} the fraction of regions highlighted in the video that are never explained by any narration step; isolates ungrounded over-highlighting that coverage alone would reward.

Each of the 12 cells receives a label in $\{1, 0.5, 0\}$ plus a required free-text rationale. Rationales make every score auditable: a label can always be traced to the evidence the annotator (human or VLM) cited, which is the property that makes the rubric portable between the two. The exact instructions given to the VLM judge (and, in paraphrased form, to the human annotators) for each of the three judgments are shown in the prompt boxes below; all three are applied independently to each of
the four axes.

\paragraph{Annotator Validation.} 
Sub-metric validation on the Non-Gold Set uses five PhD students in NLP scoring a stratified 60-video sample (20 each from low/medium/high figure complexity). For every $D_2(b)$ rubric we report Krippendorff's $\alpha$ for inter-human agreement and Cohen's $\kappa$ between the VLM judge and the human consensus (Table~\ref{tab:agreement-human}). A VLM--human $\kappa$ below the inter-human floor on any cell flags the \emph{judge} as the bottleneck and triggers prompt revision rather than rubric revision; this guardrail prevents rubric drift driven by judge weakness.

\begin{center}
\colorbox{gray!12}{%
\begin{minipage}{0.95\columnwidth}
\tiny\ttfamily
\textbf{Component Faithfulness Judge Prompt}\\[4pt]
You are a strict evaluator of whether highlighted figure components are faithful to the source figure and the paper. You judge ONE explanatory axis at a time (Inputs, Mechanism, Outputs, or Takeaway).\\[3pt]
INPUTS YOU RECEIVE:\\
1. The source figure F.\\
2. Paper context D: the figure caption, all in-text references to the figure, and the paragraphs that introduce each component shown.\\
3. The rendered video's highlighted regions that fall within this axis's scope, each as a frame with the highlight drawn on F.\\
4. The narration span spoken while each region is highlighted.\\[3pt]
WHAT TO CHECK, per (highlighted region, narration span) pair in this axis:\\
(a) LOCATION and CONNECTIVITY against F. The highlight must sit on the correct component, and the components it implies are connected must actually be connected in F by an arrow or adjacency. Penalize: pointing at the wrong box; lighting an arrow that does not exist between the named components; implying a data flow F does not show.\\
(b) ROLE ATTRIBUTION against D. The narration's claim about what the component does must match the role the paper assigns it. Penalize: calling an extractive module generative; attributing a Stage-2 operation to a Stage-1 block; any function the paper does not support.\\[3pt]
DECISION RULES:\\
- A pair is fully faithful only if BOTH (a) and (b) hold.\\
- Correct location but wrong/imprecise role, or correct role but minor connectivity error, is PARTIAL.\\
- Wrong location, fabricated connectivity, or wrong role is a FAILURE for that pair.\\
- Do not reward a pair because the correct component is merely visible; it must be the highlighted one.\\
- If you must choose between two labels, choose the lower.\\[3pt]
AXIS LABEL (aggregate the pairs you checked):\\
1.0 = all highlighted components in this axis are correct in BOTH location/connectivity and role.\\
0.5 = components are partially faithful (right place, shaky role; or right role, minor connectivity slips).\\
0.0 = components are mislocated, wrongly connected, or wrongly attributed.\\[3pt]
Return ONLY JSON:\\
\{"label": <1.0|0.5|0.0>, "pairs\_checked": [\{"region": "...", "narration": "...", "location\_ok": <true|false>, "role\_ok": <true|false>\}], "rationale": "<cite the specific evidence in F and D that justifies the label>"\}
\end{minipage}%
}
\end{center}

\begin{center}
\colorbox{gray!12}{%
\begin{minipage}{0.95\columnwidth}
\tiny\ttfamily
\textbf{Concept Coverage Judge Prompt}\\[4pt]
You measure how many gold concepts for ONE axis (Inputs, Mechanism, Outputs, or Takeaway) are BOTH highlighted AND explained in the rendered video. You are strict and require both halves.\\[3pt]
INPUTS YOU RECEIVE:\\
1. The source figure F.\\
2. The gold concepts assigned to this axis, each with: a canonical name, a list of accepted aliases, and the expected relationship(s) the gold narration states about it.\\
3. The rendered video's highlighted regions and the narration spoken for each.\\[3pt]
COVERAGE TEST, applied to each gold concept independently:\\
- HIGHLIGHTED: at the step where this concept is discussed, a region corresponding to it is visually emphasized (box, overlay, crop, zoom, pointer). Visibility alone is not enough.\\
- EXPLAINED: the narration names the concept (canonical name or a listed alias) AND states at least one correct relationship involving it (what feeds it, what it produces, or what it interacts with).\\
A concept is covered only if BOTH are true at the same step. Do NOT credit:\\
- a concept named in narration but never highlighted;\\
- a concept highlighted but never explained;\\
- a concept named with no correct relationship (a bare mention).\\[3pt]
AXIS LABEL (fraction of this axis's gold concepts that are covered):\\
1.0 = all or nearly all gold concepts for the axis are highlighted-and-explained.\\
0.5 = about half are covered, or several are highlighted with no correct relationship.\\
0.0 = few or none covered; key concepts are missing from the highlights, the narration, or both.\\[3pt]
Return ONLY JSON:\\
\{"label": <1.0|0.5|0.0>, "covered": [\{"concept": "...", "matched\_alias": "...", "relationship": "..."\}], "missing": ["..."], "rationale": "<brief reason>"\}
\end{minipage}%
}
\end{center}

\begin{center}
\colorbox{gray!12}{%
\begin{minipage}{0.95\columnwidth}
\tiny\ttfamily
\textbf{Excess Highlight Rate Judge Prompt}\\[4pt]
You measure how much highlighting within ONE axis (Inputs, Mechanism, Outputs, or Takeaway) is UNGROUNDED: highlighted but never explained by any narration step. This isolates over-highlighting that coverage alone would reward.\\[3pt]
INPUTS YOU RECEIVE:\\
1. The source figure F.\\
2. Every region the video highlights within this axis, each as a frame on F.\\
3. The full narration for this axis.\\[3pt]
PER-REGION TEST:\\
- GROUNDED: some narration step refers to this region and says something about its role or relationships.\\
- EXCESS: the region is highlighted but no step ever explains it; OR the highlight is far larger than what the narration discusses (e.g.\ an entire stage box lit while only one module is explained), in which case the unexplained surrounding area counts as excess.\\[3pt]
DIRECTION (read carefully): higher real-world excess is WORSE, but the reported label follows the same "1.0 is best" convention as the other two judgments. So map LOW excess to a HIGH label.\\[3pt]
AXIS LABEL (based on the share of highlighted regions that are unexplained):\\
1.0 = essentially every highlighted region is explained; negligible excess.\\
0.5 = a noticeable share of highlights are unexplained, or highlights are consistently oversized.\\
0.0 = much of the highlighting is ungrounded; many regions no step explains are lit.\\[3pt]
Return ONLY JSON:\\
\{"label": <1.0|0.5|0.0>, "excess\_regions": ["..."], "grounded\_regions": ["..."], "rationale": "<brief reason>"\}
\end{minipage}%
}
\end{center}

\section{Grounding Results: Full Breakdown}
\label{app:dtw_details}
 
\paragraph{Fixed-narration setting.} All grounding comparisons in Fig.~%
\ref{fig:bar_plot} and Table~\ref{tab:iou-dtw-stratified} hold the narration constant: every system is fed the same ordered human transcript $\{s^*_i\}_{i=1}^{T}$ recovered from the gold conference video, and only the grounding mechanism varies. Each system receives the same final figure image $F$ and must align the fixed narration steps to figure regions. This isolates grounding quality from narration quality, so any difference is attributable to
\emph{where} a system looks rather than \emph{what} it says.
 
\paragraph{Soft-DTW (cross-family, Fig.~\ref{fig:bar_plot}).} Soft-DTW is the single number that places \emph{every} system on one axis, including the frame-emitting generators (Veo-3.1, Code2Video, Paper2Video, CogVideoX, TEA-Image) that output no explicit regions. A calibrated VLM judge scores each (gold step, frame) pair from $1$--$10$, converted to a cost $c(i,j)=(10-\text{score})/9$ and DTW-aligned; the reported bar is $1$ minus the mean cost along the aligned path. Reading the panels:
\begin{itemize}
  \item \textbf{All / Easy.} \textsc{\sysname} leads on every backbone (All:
  Gemini $\approx0.78$, Claude $\approx0.76$, GPT $\approx0.67$). The lone
  exception is \emph{Easy}/GPT, where unconstrained VLM-Grounding is level or
  marginally ahead---on single-panel figures there is little structure for
  decomposition to exploit.
  \item \textbf{Medium $\rightarrow$ Hard.} The advantage does not merely hold,
  it \emph{widens}. On \emph{Hard}, \textsc{\sysname} stays $\approx0.66$--$0.74$
  across backbones while VLM-Grounding falls sharply (GPT to $\approx0.45$) and
  SAM Segmentation sits around $\approx0.51$--$0.57$.
  \item \textbf{End-to-end generators} are pinned to the floor in every panel
  ($\approx0.11$--$0.44$, descending Veo $\rightarrow$ Code2Video $\rightarrow$
  Paper2Video $\rightarrow$ CogVideoX $\rightarrow$ TEA-Image) and degrade
  further from Easy to Hard, because they regenerate the figure rather than
  reference it, so frame similarity to the gold collapses.
\end{itemize}
The takeaway is two-fold: \textsc{\sysname} wins on a metric that gives even the generative models a fair shot, and its margin \emph{scales with complexity}---exactly the regime the benchmark was stratified to expose.
 
\paragraph{IoU-based DTW (region-level, Table~\ref{tab:iou-dtw-stratified}).} Soft-DTW is coarse---it depends on the VLM judge and the frame-sampling rate---so the region-only IoU-based DTW serves as the precision check. It shares the \emph{same alignment, the same $[0,1]$ scale, and the same per-step macro-averaging}, differing only in the cost (per-step $1-\mathrm{F1}$ of matched vs.\ gold region sets). Because it is region-only it cannot score the video generators, which is precisely why both metrics are needed; for the same reason a soft-DTW value and an IoU-based value are \emph{not} directly comparable even when both are high. Table~\ref{tab:iou-dtw-stratified} reinforces the trend more strictly: \textsc{\sysname} wins every (tier, backbone) cell except \emph{Easy}/GPT, and the \emph{Hard} inversion is pronounced (Hard/GPT Macro-F1: \textsc{\sysname} $0.665$ vs.\ VLM-Grounding $0.524$ vs.\ SAM $0.448$). A useful nuance: SAM Segmentation often records the \emph{highest} Macro-P (precise boxes) but loses on Macro-R---accurate when it fires, but it misses the regions---whereas \textsc{\sysname} balances both and therefore wins F1, which is the quantitative form of the ``selective but complete'' claim.
 
\paragraph{Rubric axes on the extended set (Table~\ref{tab:d2b-rubric}).} On the $65$-video extended set, where no gold traces exist, grounding is scored along four explanatory axes (Inputs / Mechanism / Outputs / Takeaway), each as Component Faithfulness (CF, $\uparrow$) / Concept Coverage (CC, $\uparrow$) / Excess Highlight Rate (EH, $\downarrow$). \textsc{\sysname} posts the strongest CF and CC and the lowest EH on all four axes, with its largest improvement on \emph{Mechanism} ($.79/.74/.12$ vs.\ next-best SAM $.59/.61/.34$). EH exposes the characteristic baseline failure---VLM- and SAM-based methods over-highlight when uncertain---while diffusion/animation systems collapse on every axis.
 
\paragraph{Summary.} With narration held fixed to the human gold,
\textsc{\sysname} is the best grounder on both DTW views---the inclusive
cross-family soft-DTW and the strict region-level IoU-DTW---and the advantage grows precisely on \emph{Hard} figures, where structured decomposition and staged grounding matter most. The two metrics measure different things yet agree on the ranking, providing cross-validation for the claim, and the extended-set rubric corroborates it with selective, faithful highlighting (low EH, high CF/CC), strongest on the Mechanism axis.

\section{Complexity Stratification Details}
\label{app:complexity}
We stratify figures by \emph{structural complexity} rather than raw image size, since explanation difficulty grows with compositional structure: deeper figures require longer temporal grounding sequences, wider figures require introducing multiple components simultaneously, and multi-panel figures require cross-panel coordination and ordering.
To estimate complexity automatically, we prompt a vision--language model (Gemini~3 Pro, temperature 0, JSON-schema-constrained decoding) with the rendered figure and ask it to extract high-level structural statistics. The prompt defines a \emph{semantic unit} as a role-bearing component that would typically be introduced once during a human presentation. The model is explicitly instructed to \emph{merge rather than split} when uncertain. For example, repeated transformer layers, grouped attention heads, encoder/decoder stacks, or the internal primitives of a labeled module are treated as a single semantic unit, whereas arrows, tensor dimensions, decorative annotations, and visually repeated glyphs are ignored.
The prompt used for extraction is shown below:
\begin{quote}\small
``Count the number of high-level semantic units in this figure. A semantic unit is a role-bearing component that a presenter would typically explain once in a talk (e.g., Encoder, Decoder, Fusion Module, Retrieval Block, Input Group, Output Head). Merge visually repeated or internally repeated structures into one unit when they serve the same functional role. Do not count arrows, dimensions, repeated tensor symbols, decorative icons, or low-level sub-parts inside a labeled module unless they play independent semantic roles in the explanation.''
\end{quote}

\noindent
We extract four structural signals:
\begin{itemize}[leftmargin=*]
    \item \textbf{Component count:} Number of top-level semantic units (typically 4--10).

    \item \textbf{DAG depth:} Longest top-level dataflow path from any input unit to any output unit, ignoring internal structure within a semantic unit (typically 3--7).
    
    \item \textbf{DAG width:} Maximum number of semantic units operating in parallel at a single stage (e.g., width 1 for single pipelines, width 2 for dual-stream systems).
    
    \item \textbf{Sub-figure count:} Number of distinct panels identified through labels such as (a)/(b)/(c), explicit subtitles, or strong whitespace/frame boundaries. A single unified diagram, regardless of internal complexity, receives sub-figure count~1.
\end{itemize}

\section{Narration Qualitative Example}
\label{app:narr_qual}
We can take a look at Figure~\ref{fig:coin_case_study} where we compare  between narration generated using only the figure versus narration generated using both the paper and figure. While both narrations correctly follow the visual structure, the figure-only setting remains largely descriptive and misses the scientific motivation and intended takeaway. Adding paper context recovers the method name (COIN), explains the hard-negative design choice, and substantially improves Concept Coverage, Takeaway Recall, and External Faithfulness. Both narrations trace the same left-to-right structure faithfully, so Order Matching and Internal Faithfulness remain high in both settings. However, the figure-only narration stays largely descriptive: it lists modules and labels verbatim, never identifies the method name COIN, and fails to explain why hard negatives matter. As a result, Concept Coverage improves only partially and Takeaway Recall remains weak. In contrast, the paper-grounded narration injects the missing scientific intent. It explicitly frames the robustness motivation, explains the role of semantically equivalent paraphrases, clarifies why hard negatives provide stronger supervision than far-OOD negatives, and concludes with the intended takeaway about robustness to instruction phrasing.

\section{\sysname-Narration Team Pipeline Implementation}
\label{app:prompts}
\paragraph{Planner Agent ($\mathcal{A}_{\mathrm{plan}}$).}
The planner agent receives the retrieved evidence bundle $K_F$ together with the target architectural figure $F$ and produces a pedagogically coherent explanation order grounded in the figure structure.

\begin{tcolorbox}[colback=gray!5,colframe=black!60,title=Planner Prompt]
\footnotesize
\textbf{System Prompt:}

You are an expert conference presenter explaining a scientific architecture figure.

Given the paper context and figure image:
\begin{enumerate}
    \item Identify the major components and their relationships.
    \item Determine the best teaching order for explaining them.
    \item Follow the logical data flow of the system.
    \item Highlight the main takeaway of the figure.
\end{enumerate}

Return an ordered sequence of explanation steps. Each step should contain:
\begin{itemize}
    \item the component(s) being explained,
    \item the key claim or function,
    \item dependencies on earlier steps.
\end{itemize}

Avoid low-level visual details unless they are semantically important.
\end{tcolorbox}

\paragraph{Drafter Agent ($\mathcal{A}_{\mathrm{draft}}$).}
The drafter agent converts the structured explanation plan into natural narration aligned with the figure structure.

\begin{tcolorbox}[colback=gray!5,colframe=black!60,title=Drafter Prompt]
\footnotesize
\textbf{System Prompt:}

You are presenting a scientific paper figure to an audience at a conference.

Convert the provided explanation plan into a natural step-by-step narration.

Requirements:
\begin{itemize}
    \item Follow the provided explanation order.
    \item Explain components incrementally and coherently.
    \item Explicitly connect narration to the figure structure.
    \item Emphasize important transitions and takeaways.
    \item Keep the narration concise but pedagogically clear.
\end{itemize}

Generate one narration segment per explanation step.
\end{tcolorbox}

\subsection{Critic and Revision Prompts}
\label{app:critics}

\paragraph{Critic Agents ($\mathcal{A}_{\mathrm{crit}}^{1:4}$).}
A single narration draft often contains factual drift, missing components, redundant descriptions, or incoherent traversal order. To improve reliability, we use four specialized critic agents operating independently over the current narration draft $\{s_i\}$, the target figure $F$, and the retrieved evidence bundle $K_F$. Each critic focuses on one failure mode and returns structured feedback consisting of detected issues, supporting evidence, and suggested revisions.

\vspace{0.3em}
\noindent\textbf{Faithfulness Critic.}
Checks whether narration claims are supported by the paper evidence and figure semantics.

\begin{tcolorbox}[colback=gray!5,colframe=black!60,title=Faithfulness Critic Prompt]
\footnotesize
\textbf{System Prompt:}

You are a scientific fact-checking agent.

Given:
\begin{itemize}
    \item the paper evidence bundle,
    \item the target figure,
    \item the current narration draft,
\end{itemize}

identify statements that are unsupported, hallucinated, or inconsistent with the paper.

Focus especially on:
\begin{itemize}
    \item incorrect module functionality,
    \item unsupported causal claims,
    \item incorrect training or data-flow descriptions,
    \item terminology inconsistent with the paper.
\end{itemize}

For each issue, return:
\begin{enumerate}
    \item the problematic narration segment,
    \item why it is incorrect,
    \item the supporting paper evidence,
    \item a suggested correction.
\end{enumerate}
\end{tcolorbox}

\vspace{0.3em}
\noindent\textbf{Coverage Critic.}
Checks whether important branches, modules, or interactions are omitted.

\begin{tcolorbox}[colback=gray!5,colframe=black!60,title=Coverage Critic Prompt]
\footnotesize
\textbf{System Prompt:}

You are a scientific explanation coverage evaluator.

Given the figure structure, paper evidence, and narration draft:
\begin{itemize}
    \item identify important components or stages missing from the explanation,
    \item detect branches or interactions that are ignored,
    \item verify whether the final takeaway reflects the full pipeline.
\end{itemize}

Return:
\begin{enumerate}
    \item missing concepts or modules,
    \item why they matter,
    \item where they should appear in the narration.
\end{enumerate}
\end{tcolorbox}

\vspace{0.3em}
\noindent\textbf{Coherence Critic.}
Checks whether the narration follows a pedagogically coherent traversal order.

\begin{tcolorbox}[colback=gray!5,colframe=black!60,title=Coherence Critic Prompt]
\footnotesize
\textbf{System Prompt:}

You are evaluating whether a figure explanation follows a coherent teaching order.

Given the narration sequence and figure:
\begin{itemize}
    \item identify abrupt topic jumps,
    \item detect references to components before they are introduced,
    \item verify that the narration follows the logical data flow and stage progression of the figure.
\end{itemize}

Return:
\begin{enumerate}
    \item incoherent transitions,
    \item misplaced explanation steps,
    \item suggested reorderings.
\end{enumerate}
\end{tcolorbox}

\vspace{0.3em}
\noindent\textbf{Salience Critic.}
Checks whether the narration emphasizes the important contributions and avoids distracting redundancy.

\begin{tcolorbox}[colback=gray!5,colframe=black!60,title=Salience Critic Prompt]
\footnotesize
\textbf{System Prompt:}

You are evaluating the pedagogical salience of a scientific figure explanation.

Given the figure and narration:
\begin{itemize}
    \item identify redundant or repetitive descriptions,
    \item detect visually prominent but semantically unimportant details,
    \item verify that the explanation emphasizes the paper's main contribution and takeaway.
\end{itemize}

Return:
\begin{enumerate}
    \item unnecessary narration segments,
    \item missing emphasis points,
    \item suggested revisions to improve focus.
\end{enumerate}
\end{tcolorbox}

\paragraph{Revision Agent ($\mathcal{A}_{\mathrm{rev}}$).}
The revision agent receives the original narration draft together with feedback from all critics and produces an improved narration sequence.

\begin{tcolorbox}[colback=gray!5,colframe=black!60,title=Revision Agent Prompt]
\footnotesize
\textbf{System Prompt:}

You are revising a scientific figure explanation using critic feedback.

Given:
\begin{itemize}
    \item the current narration draft,
    \item feedback from faithfulness, coverage, coherence, and salience critics,
    \item the paper evidence and figure,
\end{itemize}

produce a revised narration sequence that:
\begin{itemize}
    \item preserves factual correctness,
    \item covers all important figure components,
    \item follows a coherent teaching order,
    \item emphasizes the main contribution and takeaway,
    \item removes unnecessary redundancy.
\end{itemize}

Generate one revised narration segment per explanation step.
\end{tcolorbox}

The revision loop continues until all critics return no major issues or a fixed iteration budget is reached. The final narration sequence is then written to the blackboard as \textsc{final}.

\section{\sysname-Grounding Pipeline Implementation Details}
\label{app:grounding-impl}

\begin{table*}[t]
\centering
\footnotesize
\setlength{\tabcolsep}{5pt}
\renewcommand{\arraystretch}{1.15}

\begin{tabularx}{\textwidth}{
@{}>{\raggedright\arraybackslash}p{2.8cm}
>{\raggedright\arraybackslash}p{3.0cm}
>{\raggedright\arraybackslash}X
>{\raggedright\arraybackslash}p{3.2cm}
>{\centering\arraybackslash}p{0.7cm}@{}
}
\toprule
\textbf{Stage / Component} & \textbf{Model(s)} & \textbf{Key settings} & \textbf{License / Terms} & \textbf{HF} \\
\midrule

Narration \& grounding backbones 
& Gemini-3.1-Pro, GPT-5, Claude Sonnet 
& Multi-agent narration and grounding generation 
& Google / OpenAI / Anthropic ToS 
& No \\

Complexity tiering 
& Gemini-3-Pro 
& Temperature $=0$; JSON-schema constrained output 
& Google API ToS 
& No \\

OCR module ($A_{\text{ocr}}$) 
& Azure Document Intelligence 
& Paragraph-level OCR boxes 
& Microsoft Azure ToS 
& No \\

Visual detector ($A_{\text{vis}}$) 
& SAM3 (cloud endpoint) 
& Detection threshold $=0.25$; sliding-window inference 
& Endpoint provider terms 
& No \\

Optional detector 
& GroundingDINO (Swin-T OGC) 
& Optional fusion; default pipeline uses SAM3 only 
& Apache 2.0 
& Yes \\

Embedding / clustering 
& DINOv2 
& Glyph cosine $\geq 0.88$; cluster cosine $\geq 0.92$; distance $\leq 2.5$ 
& Apache 2.0 
& Yes \\

TTS 
& Inworld Mini (default), OpenAI TTS (optional) 
& Segment duration $d_i = \lfloor 1000\tau_i \rfloor$ 
& Inworld / OpenAI ToS 
& No \\

Video renderer 
& Manim Community Edition + ffmpeg 
& Figure $F$ used as static background; rendered highlight overlays 
& MIT 
& n/a \\

\midrule

Narration judges (evaluation) 
& Gemini-2.5-Flash/Pro, GPT-4o, Claude-3.7-Sonnet, DeepSeek-R1 
& Shared prompts and evidence across all judges 
& Provider ToS; DeepSeek-R1 weights under MIT 
& Partial \\

Soft-DTW judge (evaluation) 
& Gemini-3-Flash (selected), GPT 
& Strict 1--10 rubric-based alignment scoring 
& Provider ToS 
& No \\

Baselines 
& Veo 3.1, CogVideoX, Code2Video, TheoremExplainAgent, Paper2Video 
& Adapted using figure $F$ and gold narration $\{s^\ast\}$ 
& Mixed licenses; see provider/repository terms 
& Partial \\

\bottomrule
\end{tabularx}

\caption{
Models, hyperparameters, and licenses used in \sysname{} and evaluation. API-hosted models are governed by provider terms of service (ToS) rather than open-source licenses. ``HF'' indicates whether public HuggingFace weights/checkpoints are available.
}
\label{tab:models_licenses}

\end{table*}

This appendix records the detectors, thresholds, and serialization formats abstracted away in Sec.~\ref{sec:experiments}. The pipeline executes $\Phi \rightarrow \mathcal{G} \rightarrow \textsc{Render}$ over a raster figure $F$ and a fixed narration $\{s_i\}_{i=1}^{T}$, producing the intermediate inventory \texttt{detections.json}, the region whitelist \texttt{regions.json}, and the grounded script \texttt{script.json}; the final output is the script $\mathcal{S}=\{(\ell_i,\mathbf{r}_i,d_i)\}_{i=1}^{T}$ with $\ell_i\equiv s_i$ and optional video $V$ with per-step audio duration $\tau_i$.

\paragraph{Figure decomposition ($\Phi$).}
The text layer uses Azure Document Intelligence for paragraph-level OCR, yielding text detections $\mathcal{T}=\{\tau_j\}$ with boxes and transcripts. The visual layer uses a SAM3 \citep{carion2025sam} open-vocabulary detector (cloud endpoint) to segment icons, blocks, arrows, and illustrations into $\mathcal{V}=\{\nu_k\}$ with axis-aligned boxes and optional polygons; GroundingDINO \citep{liu2024grounding} may be fused in a hybrid mode, though the default deployment uses SAM3 alone with arrow instances taking precedence over overlapping legacy boxes. Before clustering, $\Phi$ runs a deterministic cleanup cascade: (i) suppress visuals that duplicate OCR (IoU $\geq 0.7$ against raw text boxes); (ii) remove nested container detections and regions covering $>50\%$ of the image; (iii) recover repeated glyphs via shape-template matching, retaining candidates only if DINOv2 embedding cosine similarity $\geq 0.88$; and (iv) merge tiny overlapping boxes (area $\leq 0.2\%$ of the slide, IoU $\geq 0.1$). Clustering then builds a graph over $\mathcal{T}\cup\mathcal{V}$ with edges when DINOv2 cosine similarity $\geq 0.92$ and normalized spatial distance $\leq 2.5$; connected components of size $\geq 2$ become clusters $\mathcal{C}=\{c_m\}$, recorded with merged membership in \texttt{detections.json}. A builder assigns stable identifiers $t_j$, $v_k$, $c_m$ (each with pixel and normalized boxes, display strings, and optional polygons) and writes $\mathcal{U}=\{t_j\}\cup\{v_k\}\cup\{c_m\}$ to \texttt{regions.json}.

\paragraph{Narration-to-region grounding ($\mathcal{G}$).}
For each step, $\mathcal{G}$ constructs a candidate packet from $\mathcal{U}$ capped at $60$ entries for context limits, enriching visual and cluster candidates with nearest OCR snippets. A multimodal model (default: Gemini Flash) receives $F$, the ordered narration $\{s_i\}$, and the packet, and selects region and cluster IDs for each step. Pass-1 explanation planning from the slide alone is skipped when $\{s_i\}$ is supplied. A schema validator enforces the JSON structure and filters any ID outside $\mathcal{U}$; empty or under-specified steps are repaired against region summaries, and optional post-processing injects connector-arrow metadata when layout implies directed flow between consecutive $\mathbf{r}_i$.

\paragraph{Renderer.}
For each non-empty $\ell_i$, a TTS backend (default: Inworld mini; OpenAI speech optional) synthesizes \texttt{manim\_narration\_part\_$i$.mp3}, probes duration $\tau_i$, and sets \texttt{audio\_duration\_s}$_i=\tau_i$ and $d_i=\lfloor 1000\,\tau_i\rfloor$; concatenated audio defines the master track. A VLM generates Manim Community Edition code that keeps $F$ as a persistent background, schedules one annotation phase per step anchored to $\mathbf{r}_i$ (text recolor, lifted visual crops, conservative callouts) with \texttt{step\_duration\_s} tied to $\tau_i$, and omits on-screen subtitles. The scene is rendered to a silent MP4 and muxed with the narration via \texttt{ffmpeg}. The same script can alternatively emit per-step JPEG highlights, interactive web overlays, or quiz items consuming $(\ell_i,\mathbf{r}_i)$, all sharing $\mathcal{G}$'s output without altering $\{s_i\}$.

\newcommand{\benchname}{\textsc{FigTalk}\xspace}

\section{Narration Ablations}
\label{sec:narration-ablation}

While the grounding ablations in Table~\ref{tab:narration_ablation2} isolate the contribution of the decomposition pipeline, the gains in $D_1$ also depend strongly on the upstream narration generator $\mathcal{N}$. In particular, \sysname introduces two narration-specific design choices absent from standard figure captioning or slide narration systems: (i) retrieval-augmented evidence assembly through the figure-specific context bundle $K_F$, and (ii) the critic--reviser refinement loop enforcing faithfulness, coverage, coherence, and salience constraints.

To isolate their contribution, we perform narration-side ablations using the Paper+Figure regime with \textsc{Gemini-3.1-Pro} as the backbone, since it is the weakest backbone in the main experiments and therefore the setting where pipeline structure should matter most. We evaluate four variants: (a) the full narration pipeline, (b) without retrieval ($-K_F$), where the model receives only the figure and caption, (c) without the critic--reviser loop, and (d) with a weakened evidence bundle restricted to the caption and abstract only.





\paragraph{Retrieval is responsible for the largest gains in contextual explanation quality.}
Removing the figure-specific evidence bundle $K_F$ causes the largest drop on Concept Coverage, Takeaway Recall, and External Faithfulness. This confirms that architectural figures alone are insufficient for recovering the intended contribution of the paper. Without retrieval, the narrator often falls back to generic architectural descriptions (“encoder,” “attention block,” “feature extractor”) while missing the paper-specific novelty and rationale behind those components.

\paragraph{The critic--reviser loop primarily improves factual consistency and traversal coherence.}
Disabling refinement substantially reduces Internal Faithfulness and Order Matching. In practice, single-pass drafts frequently introduce unsupported causal claims, skip intermediate components, or revisit earlier modules in inconsistent orderings. The refinement loop reduces these errors by explicitly enforcing coverage and traversal constraints before grounding begins.

\paragraph{Weak evidence bundles produce superficially fluent but incomplete narrations.}
Restricting the evidence bundle to only the caption and abstract preserves fluency but reduces deeper explanatory quality. Narrations generated under this setting often identify the high-level purpose of the architecture but omit intermediate module interactions and detailed rationales that appear only in the method section or in-text figure references.

\subsection{Discussion}
\label{sec:discussion}

\paragraph{Separating narration from grounding improves diagnosability.}
A central design choice in \sysname is the explicit factorization of figure explanation into two stages: narration generation and narration-conditioned grounding. Existing end-to-end systems often conflate these capabilities, making it difficult to determine whether failures originate from incorrect explanations or incorrect visual alignment. By evaluating $D_1$ and $D_2$ independently, \sysname exposes complementary failure modes that would otherwise remain hidden inside a single aggregate score.

\paragraph{Architecture figures require paper-grounded reasoning rather than pure visual understanding.}
The strongest improvements consistently appear on metrics tied to explanation intent rather than raw localization accuracy. Concept Coverage, Takeaway Recall, and Mechanism grounding all depend on understanding why a component exists and how it contributes to the paper’s contribution. These signals are rarely recoverable from pixels alone. The results therefore suggest that figure explanation is fundamentally a multimodal reasoning task grounded jointly in document context and visual structure.

\paragraph{Over-highlighting is a major failure mode of direct VLM grounding.}
Across both $D_2(a)$ and $D_2(b)$, VLM-Grounding and SAM-based baselines frequently highlight large or weakly related regions when uncertain. While this behavior can preserve recall, it reduces interpretability and makes the walkthrough visually noisy. Excess Highlight Rate exposes this trade-off directly: systems without structured decomposition hedge uncertainty by highlighting too much of the figure at once.

\paragraph{Re-rendering systems optimize visual fluency at the expense of fidelity.}
Diffusion-based and Manim-style systems occasionally produce visually appealing animations, but they drift from the source figure structurally. Human annotators repeatedly preferred grounded overlays on the original figure over regenerated visuals because scientific communication requires exact component-level correspondence rather than stylistic plausibility. This explains why diffusion systems perform especially poorly on Component Faithfulness despite generating polished videos.

\paragraph{The current pipeline is intentionally architecture-figure specific.}
\sysname currently focuses on architecture-style scientific figures where meaning is carried through modules, arrows, and sequential dataflow. Results panels, plots, and statistical charts require fundamentally different decomposition and narration strategies centered around trends, numerical comparisons, and uncertainty communication. Extending the framework to these figure families remains important future work.

\subsection{Inference Cost and Latency Analysis}
\label{sec:cost-analysis}

\paragraph{\sysname trades additional inference stages for substantially improved grounding fidelity.}
Unlike single-pass VLM grounding systems, \sysname performs retrieval, staged decomposition, grounding, and rendering sequentially. This increases end-to-end latency but produces significantly more faithful and interpretable explanations.

\begin{table*}[t]
\centering
\small
\setlength{\tabcolsep}{5pt}
\begin{tabular}{lcccccc}
\toprule
Method & Narration & Grounding & Rendering & Time & API Cost & References Original $F$? \\
\midrule
VLM-Grounding & single-pass & direct VLM & overlay & $\sim$18s & low & \cmark \\
SAM Seg + BBox & retrieval-lite & SAM + VLM & overlay & $\sim$32s & low--mid & \cmark \\
\sysname & retrieval + critics & staged grounding & TTS-aligned & $\sim$72s & mid--high & \cmark \\
Paper2Video & slide narration & coarse cursor & slide video & $\sim$45s & mid & \cmark  \\
TEA & prompt planning & re-rendering & Manim & $\sim$2--5 min & high & \xmark \\
Code2Video & prompt planning & re-rendering & Manim & $\sim$2-3 min & high & \xmark \\
Veo-3.1 & prompt-only & implicit & diffusion video & $\sim$30s & very high & \xmark \\
CogVideoX & prompt-only & implicit & diffusion video & $\sim$38s & very high & \xmark \\
\bottomrule
\end{tabular}
\caption{Approximate runtime and deployment characteristics per figure. Runtime includes narration, grounding, and rendering. Diffusion and re-rendering systems are substantially slower while also failing to preserve direct correspondence to the source figure.}
\label{tab:cost-analysis}
\end{table*}

\paragraph{Most of \sysname's overhead comes from narration refinement rather than grounding.}
Profiling shows that the critic--reviser loop dominates runtime, particularly when multiple refinement iterations are required for difficult figures. In contrast, figure decomposition and grounding are comparatively lightweight once the region pool $U$ is constructed. This suggests that future efficiency gains are more likely to come from improving narration planning than from optimizing grounding.

\paragraph{Diffusion and re-rendering systems are substantially slower despite weaker grounding quality.}
Manim-based and diffusion-based systems require either iterative code generation or autoregressive video synthesis, both of which introduce large latency overheads. Despite these costs, their outputs remain poorly aligned with the original figure because they regenerate rather than directly reference $F$.

\paragraph{The staged pipeline also improves controllability and auditability.}
Although \sysname introduces additional stages, the intermediate outputs ($K_F$, region pools, grounding traces, and narration steps) are fully inspectable and debuggable. This makes failures diagnosable in a way end-to-end video generation systems are not, which is especially important for scientific and educational applications.

\section{Related Work}
\label{sec:related-work}

Prior work related to \sysname falls into four broad categories: scientific figure understanding, document-to-presentation systems, grounded visual explanation systems, and video generation models.

\paragraph{Scientific figure understanding and grounding.}
Early work focused primarily on static understanding tasks such as figure captioning and question answering. Datasets such as SciCap~\citep{hsu-etal-2021-scicap-generating} and FigureQA~\citep{EbrahimiKahou2017FigureQAAA} evaluate whether models can describe or reason about figures, while ChartQA~\citep{masry-etal-2022-chartqa} studies chart-specific visual reasoning. More recent grounding-oriented approaches such as TextHawk2 and HiVG improve fine-grained OCR and referring-expression localization, but they operate on isolated queries rather than constructing temporally ordered walkthroughs.

\paragraph{Document-to-presentation and paper-to-video systems.}
Systems such as Doc2PPT~\citep{fu2022doc2pptautomaticpresentationslides}, D2S~\citep{sun-etal-2021-d2s}, PPTAgent~\citep{zheng-etal-2025-pptagent}, and Paper2Video~\citep{zhu2025paper2videoautomaticvideogeneration} generate slides or narrated presentations from scientific papers. However, these systems typically treat figures as opaque visual assets and rely on coarse cursor motion or slide-level narration rather than explicit region-level grounding.

\paragraph{Animated explanation and theorem visualization systems.}
Recent systems such as TheoremExplainAgent~\citep{ku-etal-2025-theoremexplainagent} and Code2Video~\citep{code2video} generate educational animations using programmatic rendering frameworks such as Manim. While these systems produce sequential visual explanations, they are designed for abstract theorem or code explanation rather than faithful grounding to a fixed source figure.

\paragraph{Diffusion-based video generation.}
General text-to-video systems such as Veo and CogVideoX~\citep{Yang2024CogVideoXTD} can generate visually rich animations from prompts, but they lack explicit mechanisms for preserving scientific figure fidelity or grounding narration to exact figure components.

\begin{table*}[t]
\centering
\tiny
\setlength{\tabcolsep}{4pt}
\begin{tabular}{lcccccc}
\toprule
Method Family & Paper-grounded & Region-level grounding & Sequential walkthrough & References original figure & Narration evaluation & Grounding evaluation \\
\midrule
SciCap / Figure Captioning & partial & \xmark & \xmark & \cmark & partial & \xmark \\
ChartQA / FigureQA & \xmark & partial & \xmark & \cmark & \xmark & partial \\
Doc2PPT / D2S / PPTAgent & \cmark & \xmark & partial & partial & partial & \xmark \\
Paper2Video & \cmark & coarse & partial & partial & partial & partial \\
TheoremExplainAgent & partial & \xmark & \cmark & \xmark & \xmark & \xmark \\
Code2Video & partial & \xmark & \cmark & \xmark & \xmark & \xmark \\
Diffusion Video Models & \xmark & implicit & partial & \xmark & \xmark & \xmark \\
\midrule
\sysname (ours) & \cmark & \cmark & \cmark & \cmark & \cmark & \cmark \\
\bottomrule
\end{tabular}
\caption{Comparison with related work. Existing methods typically optimize only subsets of the figure-explanation problem, whereas \sysname jointly addresses paper grounding, sequential narration, region-level alignment, and figure-faithful rendering.}
\label{tab:related-work}
\end{table*}

\section{Scope: Architecture Figures vs.\ Charts/Results}
\label{app:scope}
\sysname{} deliberately targets \emph{architecture-style} figures---pipeline diagrams, neural architectures, module-interaction graphs, and system overviews---where meaning is carried by spatial composition and ordered information flow, so a temporally-ordered walkthrough is well-defined. We \emph{exclude} results panels (charts, ablation tables) by design, for two reasons. (i) The reading objective is different: a result panel is explained by \emph{quoting values and trends}, not by tracing a DAG of components, so the ``pedagogical reading order'' that drives $\mathcal{N}$ is undefined. (ii) Chart grounding is a distinct, already-active problem with its own benchmarks (e.g., ChartQA \citep{masry-etal-2022-chartqa}, FigureQA \citep{EbrahimiKahou2017FigureQAAA}; a chart-specific extractor would be the right $\Phi$ for that regime. Architecturally, only $\Phi$ is figure-type-specific: $\mathcal{N}$, $G$, and the renderer are agnostic to how the region pool $U$ was produced. Extending \sysname{} to charts is therefore a matter of swapping in a chart-aware decomposer (data-cell / axis / legend extraction) and a value-tracing narration objective---a clean direction we leave to future work and now state explicitly in the Limitations.

\section{\sysname{} Implementation Details}

\subsection{Retrieval Agent}
\label{app:retrieval}
$\mathcal{A}_{\mathrm{ret}}$ indexes $D$ at paragraph granularity with dense embeddings and, for each $F$, assembles $K_F$ from (a)~the caption and all in-text references to $F$ by label, (b)~the top-$k$ paragraphs most similar to the caption and to OCR strings from $F$, and (c)~abstract and contribution sentences from the introduction. A cross-encoder reranks $K_F$ and a deduplication pass removes near-duplicates, yielding a compact, figure-specific
evidence set.


\begin{table*}[!t]
\centering
\begin{tabular}{lccccc}
\toprule
Configuration & OM & IF & CC & TR & EF \\
\midrule
\sysname (full) & \textbf{0.76} & \textbf{0.83} & \textbf{0.84} & \textbf{0.81} & \textbf{0.82} \\

\midrule
\multicolumn{6}{c}{\textit{Remove narration capabilities}} \\

\hspace{2mm} w/o retrieval $K_F$ & 0.70 & 0.31 & 0.42 & 0.46 & 0.29 \\

\hspace{2mm} w/o planner & 0.61 & 0.79 & 0.68 & 0.63 & 0.71 \\

\hspace{2mm} w/o critic-reviser & 0.69 & 0.71 & 0.61 & 0.58 & 0.63 \\

\hspace{2mm} caption-only evidence & 0.67 & 0.54 & 0.49 & 0.41 & 0.46 \\

\hspace{2mm} figure-only & 0.73 & 0.80 & 0.60 & 0.43 & 0.28 \\

\bottomrule
\end{tabular}
\caption{Narration-side ablations. Retrieval of paper
evidence $K_F$ is the primary driver of scientific
faithfulness and takeaway quality, while the planner
and critic-reviser stages mainly improve pedagogical
ordering and coherence.}
\label{tab:narration_ablation2}
\end{table*}

\subsection{Detection and Segmentation}
\label{app:detect}
$\mathcal{A}_{\mathrm{ocr}}$ produces text regions $\mathcal{T}=\{\tau_j\}$ with bounding boxes and transcripts. $\mathcal{A}_{\mathrm{vis}}$ detects visual components $\mathcal{V}=\{\nu_k\}$ via open-vocabulary detection and segmentation, with sliding-window detection to recover small objects. On overlap, segmentation boundaries are preferred for tighter highlights, and arrows are handled so directional flow is preserved. $\mathcal{A}_{\mathrm{clean}}$ then removes duplicates, suppresses oversized containers, merges fragments, and recovers repeated symbols via shape matching, all through a fixed
rule-based pipeline for determinism.

\subsection{Clustering and Region Building}
\label{app:builder}
$\mathcal{A}_{\mathrm{clus}}$ computes visual embeddings over $\mathcal{T}\cup\mathcal{V}$
and links detections that are visually similar, spatially close, and semantically
    compatible (text regions additionally require transcript similarity), forming clusters
$\mathcal{C}=\{c_m\}$; a validation step removes groups containing unrelated elements.
$\mathcal{A}_{\mathrm{reg}}$ assigns stable IDs, bounding boxes, and display names, yielding
$\mathcal{U} = \mathcal{T} \cup \mathcal{V} \cup \mathcal{C}$. Arrows are handled separately
at render time. Optionally, the builder partitions the figure into coarse spatial zones
(\eg, left/middle/right) to disambiguate visually similar elements.

\subsection{Grounding, Validation, and Rendering}
\label{app:grounding}
$\mathcal{A}_{\mathrm{pack}}$ builds a small candidate packet per step from $\mathcal{U}$,
enriching visual elements with nearby OCR text and (if available) tagging candidates with
their spatial zone. $\mathcal{A}_{\mathrm{sel}}$ maps each $s_i$ to one or more region IDs;
text/cluster regions are shown as highlighted boxes, visual regions as cropped views.
$\mathcal{A}_{\mathrm{val}}$ removes regions outside $\mathcal{U}$ and repairs incomplete
bindings using region summaries, producing the grounded script:
\begin{equation}
  \label{eq:grounded-step}
  \text{step}_i =
  \bigl(\texttt{step\_id}_i,\; \ell_i,\; \mathbf{r}_i,\; d_i\bigr),
  \qquad \ell_i = s_i,
\end{equation}
where $\mathbf{r}_i$ is the selected highlight region and $d_i$ the initial display
duration. $\mathcal{A}_{\mathrm{rend}}$ synthesizes speech for each step and sets each
highlight's on-screen duration from the measured audio length, then overlays highlights on
the original figure during narration. The same grounded script also supports interactive
overlays, step-wise image highlights, or quiz interfaces.

\subsection{Aggregation}

We report each metric separately rather than collapsing them into a single score because the metrics capture distinct failure modes. For example, a narration may remain faithful to the paper while presenting concepts in a poor pedagogical order, or it may preserve ordering while failing to communicate the paper's central takeaway.

When a single aggregate score is required, we compute the unweighted macro-average:
\[
\textsc{NarrationScore}
=
\frac{1}{|\mathcal{M}|}
\sum_{m \in \mathcal{M}} m(Y),
\]
where $\mathcal{M}$ contains the applicable metrics among Order Matching, Internal Faithfulness, Concept Coverage, Takeaway Recall, and External Faithfulness.

\section{Human Evaluation Details}
\label{app:human-eval}

\paragraph{Systems and confound control.}
All systems receive the same input paper $D$ and figure $F$, and---where the system supports external narration---the same gold transcript $\{s^*_i\}$ used in the $D_2(a)$ protocol. Output videos are length-normalized to within $\pm10\%$ of one another to remove duration as a confound, and audio is rendered with the same TTS voice across all systems where the system does not produce its own audio.

\paragraph{Criteria and rubric.}
Annotators rank variants ($1=$ best) on six criteria across two axes plus an overall preference. \emph{Narration quality}: (N1) faithfulness,
(N2) comprehensibility, (N3) coherence. \emph{Grounding quality}: (G1) absence of excess highlighting, (G2) absence of redundant highlighting across steps, (G3) correct step-by-step grounding. (O) Overall preference.

\paragraph{Study I --- extended results.}
The faithfulness gap is the largest single effect (N1 rank $1.2$ for \textsc{N-Step (Paper+Figure)} vs.\ $2.9$ for the figure-only step narration), confirming that paper context supplies claims the figure alone cannot. The summary variant trails on comprehensibility and coherence because it compresses multi-step reasoning into a single pass. Both grounding axes are similar across narration sources, as expected given the shared \textsc{\sysname} pipeline. Inter-rater agreement (Kendall's $W$) is $0.66$ overall.

\paragraph{Study II --- extended results.}
With narration shared, the narration criteria (N1--N3) stay close while the grounding criteria (G1--G3) separate the variants. \textsc{\sysname} achieves the best step-by-step grounding (G3 rank $1.3$ vs.\ $2.6$ for \textsc{VLM-Grounding}) and the least redundant highlighting. \textsc{Code2Video} ranks last on grounding, frequently over-highlighting regions unrelated to the current step. Inter-rater agreement is higher on this grounding-dominated comparison
(Kendall's $W=0.72$ overall).

\begin{figure*}[t]
  \centering
  \includegraphics[width=0.98\textwidth]{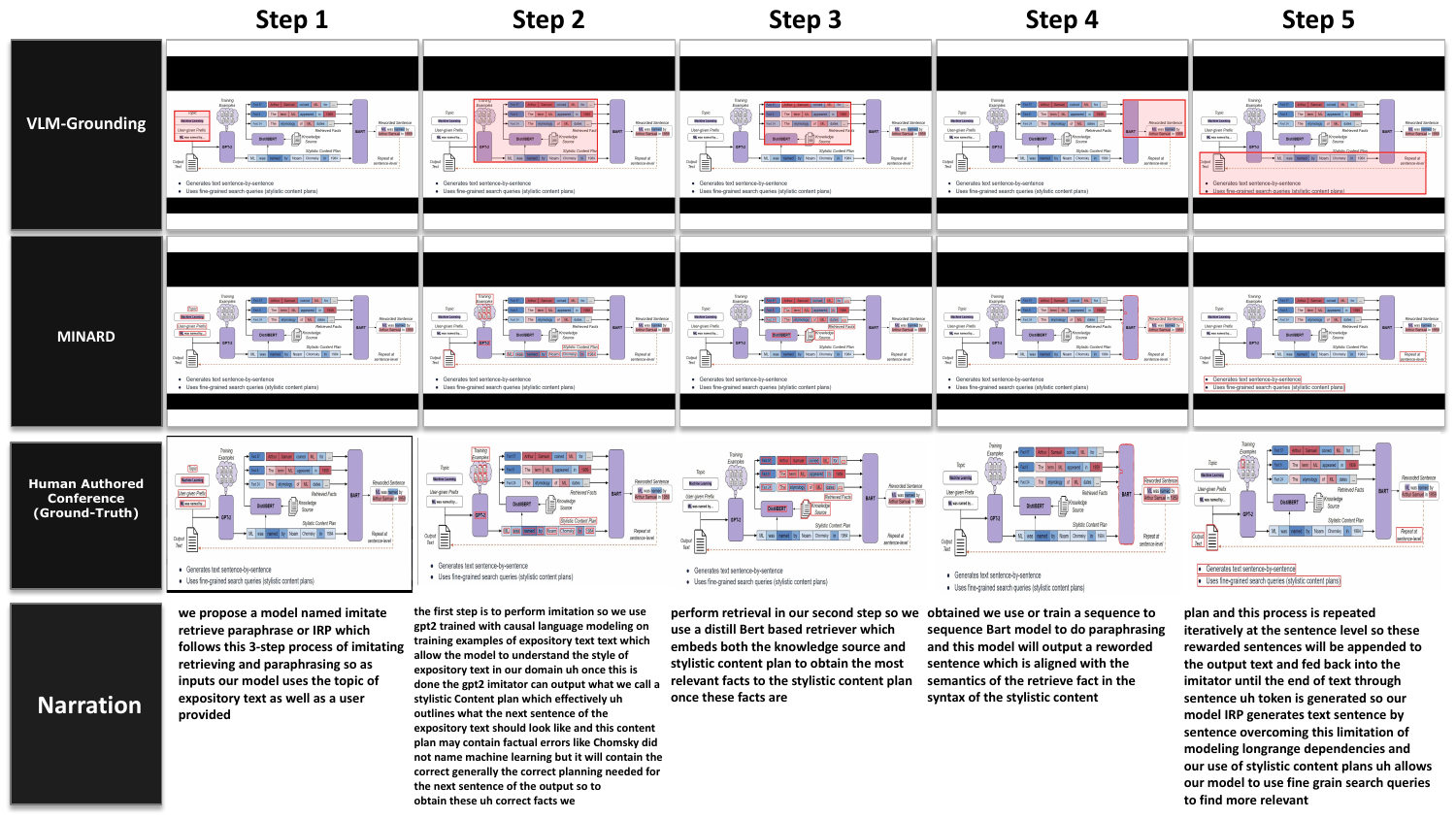}
  \caption{Outline of \sysname with a running example of the NapSS architecture~\cite{lu-etal-2023-napss}, a two-stage narrative-prompting and sentence-matching pipeline for plain-language summarization. Stage~1 trains a BERT extractive summarizer on sentence-matched ABS/PLS pairs; Stage~2 combines a Stanza-based narrative prompt with a BERT extractive summary and feeds both to BART to generate the final PLS.}
  \label{fig:example_input}
\end{figure*}

\begin{table}[h]
\centering
\small
\caption{Inter-rater agreement (Kendall's $W$, $\uparrow$) per criterion, per
study.}
\label{tab:agreement}
\setlength{\tabcolsep}{5pt}
\begin{tabular}{lccccccc}
\toprule
& N1 & N2 & N3 & G1 & G2 & G3 & O \\
\midrule
Study I  ($W$) & 0.68 & 0.61 & 0.63 & 0.59 & 0.57 & 0.60 & 0.66 \\
Study II ($W$) & 0.58 & 0.55 & 0.56 & 0.73 & 0.70 & 0.76 & 0.72 \\
\bottomrule
\end{tabular}
\label{tab:agreement-ranking}
\end{table}

\section{Grounding Baselines: Family Details}
\label{app:grounding-baselines}

\subsection{Baseline Implementation Details}
\label{app:baseline-implementations}

All grounding baselines in D2 operate under the same evaluation setting: the narration is fixed to the gold transcript $\{s_i^\ast\}_{i=1}^{T}$ recovered from the reference conference video, and only the grounding mechanism varies. Each system receives the same final figure image $F$ and must generate a grounding trace aligning narration steps to visual regions of the figure. This isolates grounding quality from narration quality and ensures that performance differences arise from the grounding pipeline rather than differences in generated explanations.

\paragraph{VLM-Grounding.}
The VLM-Grounding baseline evaluates whether a frontier vision-language model can directly localize narration steps on a scientific figure without any explicit figure decomposition or intermediate region inventory. Unlike \sysname{}, which separates figure perception from narration grounding through the Perception Team $\Phi$ and the Grounding Team $G$, this baseline collapses perception, semantic interpretation, and localization into a single unconstrained prediction problem.

The model receives the final figure image $F$, the image width and height in pixels, and the ordered transcript steps $\{s_i^\ast\}_{i=1}^{T}$. For each narration step, the model predicts exactly one coarse bounding box directly in image pixel space. Bounding boxes are represented as
\[
[left_x, top_y, right_x, bottom_y],
\]
with coordinates defined relative to the original image resolution and referenced from the top-left image origin. If a narration step is not visibly grounded in the figure, the model may return \texttt{null}.

\begin{center}
\colorbox{gray!12}{\begin{minipage}{0.96\linewidth}\small\ttfamily
\textbf{System Prompt}

You ground explanation steps to figure regions in an image.

Return ONLY valid JSON with schema:

\{"boxes": [{"step\_number": 1,
"line": "...",
"bbox": [left\_x, top\_y, right\_x, bottom\_y] | null}]\}

Bounding boxes must use absolute image pixel coordinates referenced from the top-left image origin.

Return exactly one item per narration step.

Prefer tight boxes around the relevant figure region. Do not return full-image boxes unless the entire figure is required for the narration step.
\end{minipage}}
\end{center}

\begin{center}
\colorbox{gray!12}{\begin{minipage}{0.96\linewidth}\small\ttfamily
\textbf{User Prompt}

Image width: W pixels.

Image height: H pixels.

Coordinate contract:
Return bounding boxes as left x, top y, right x, bottom y using the exact original image dimensions. Do not return normalized coordinates, percentages, or coordinates from resized crops.

Ground each narration step to one coarse bounding box around the visible figure region required for that step.

If a narration step is not visibly supported by the figure, return null.

Steps:
The ordered transcript narration steps.
\end{minipage}}
\end{center}

Unlike \sysname{}, the model is not given OCR outputs, segmentation masks, semantic clusters, candidate-region identifiers, or any constrained grounding inventory. Thus, there is no equivalent of the Perception Team $\Phi$, no region whitelist $U$, and no validator-based repair stage. The model must independently infer the relevant visual region while simultaneously predicting coordinates.

To enable evaluation under the same region-level metrics as \sysname{}, predicted VLM boxes are mapped post-hoc to backend-extracted candidate regions. A backend region is considered selected if at least $50\%$ of its area lies inside the predicted VLM box.

\paragraph{SAM Segmentation + BBox.}
The SAM Segmentation + BBox baseline evaluates whether backend-extracted candidate regions alone are sufficient for narration grounding. Unlike VLM-Grounding, the model does not directly predict coordinates. Instead, grounding is formulated as constrained selection over a pre-extracted inventory of candidate regions.

The backend extraction stage first processes the figure independently of the narration. Text regions are extracted using Azure Document Intelligence OCR, retaining paragraph-level boxes to reduce token-level fragmentation. Visual candidates are extracted using a SAM3 segmentation endpoint hosted on RunPod using the vocabulary:
\begin{quote}
\small
\texttt{visual, icon, graphic, visual component, illustration, graphic component, image}
\end{quote}

The detector uses a detection threshold of $0.25$ and a mask threshold of $0.7$. Extracted regions include architecture blocks, arrows, icons, glyphs, diagrams, and other non-textual components. Each candidate region contains a region identifier, region type, bounding box, and optional OCR-associated text.
Unlike \sysname{}, however, the extracted candidates remain flat and unstructured. The baseline does not perform semantic clustering, narration-aware grouping, duplicate suppression, connector grouping, or validator-based repair.
Given the figure image, extracted candidate inventory, and transcript steps $\{s_i^\ast\}_{i=1}^{T}$, the model selects zero, one, or multiple region identifiers for each narration step.

\begin{center}
\colorbox{gray!12}{\begin{minipage}{0.96\linewidth}\small\ttfamily
\textbf{System Prompt}

You ground transcript steps to a provided set of extracted figure regions in an image.

Return ONLY valid JSON with schema:

\{"boxes": [{"step\_number": 1,
"line": "...",
"region\_ids": [1,2]}]\}

Use only region ids from the provided candidate list.

Return exactly one item per narration step.

Each narration step may map to zero, one, or multiple region ids.

Prefer the minimal set of region ids necessary to support the narration step.

If a narration step is not visibly supported, return an empty list.
\end{minipage}}
\end{center}

\begin{center}
\colorbox{gray!12}{\begin{minipage}{0.96\linewidth}\small\ttfamily
\textbf{User Prompt}

Image width: W pixels.

Image height: H pixels.

Candidate extracted regions:
JSON list containing region id, region type, bounding box, and optional OCR text.

Select the one or more extracted region ids supporting each narration step.

Steps:
The ordered transcript narration steps.
\end{minipage}}
\end{center}

\paragraph{Implementation Details of PaperTalker.}
This baseline follows the subtitle-builder and cursor-builder formulation proposed in PaperTalker~\cite{zhu2025paper2videoautomaticvideogeneration}. Unlike \sysname{}, the pipeline does not construct an explicit semantic region inventory or perform constrained narration-aware grounding over extracted figure components.

Given the final scientific figure $F$ and the gold narration transcript $\{s_i^\ast\}_{i=1}^{T}$, the system first converts each narration sentence into a visual-focus instruction describing where a presenter cursor should point while the sentence is spoken.

\begin{center}
\colorbox{gray!12}{\begin{minipage}{0.96\linewidth}\small\ttfamily
\textbf{Visual-Focus Generation Prompt}

You are given a scientific figure and a narration sentence from an explanation video.

Generate a short visual-focus instruction describing where a presenter cursor should point while speaking this sentence.

The instruction should refer only to visible regions in the figure and should be concise, spatially grounded, and presentation-oriented.
\end{minipage}}
\end{center}

\begin{center}
\colorbox{gray!12}{\begin{minipage}{0.96\linewidth}\small\ttfamily
\textbf{User Prompt}

Figure:
Scientific figure image.

Narration sentence:
Current explanation sentence.

Return a short cursor-guidance instruction describing which visible region should be emphasized while the sentence is spoken.
\end{minipage}}
\end{center}

The generated visual-focus instruction is then grounded into a cursor location using a GUI-grounding model similar to UI-TARS.

\begin{center}
\colorbox{gray!12}{\begin{minipage}{0.96\linewidth}\small\ttfamily
\textbf{Cursor Grounding Prompt}

You are a cursor-grounding model for scientific figures.

Given a figure image and a visual-focus instruction, predict the screen location where a presenter cursor should point.

Return only JSON with schema:

\{"x": number, "y": number\}

Coordinates must be normalized to the range [0,1].
\end{minipage}}
\end{center}

\begin{center}
\colorbox{gray!12}{\begin{minipage}{0.96\linewidth}\small\ttfamily
\textbf{User Prompt}

Visual-focus instruction:
Current cursor guidance sentence.

Predict the cursor location corresponding to the visually relevant figure region.
\end{minipage}}
\end{center}

Unlike \sysname{}, which grounds narration directly to semantically extracted figure regions and grouped interaction structures, this baseline grounds only to cursor points inferred from free-form visual-focus prompts. There is no equivalent of the Perception Team $\Phi$, no narration-aware packet construction, and no validator-based repair stage.
To enable direct comparison with \sysname-Grounding, predicted cursor points are mapped back to backend-extracted candidate regions using spatial containment and neighborhood overlap.

\paragraph{Implementation Details of Code2Video \citep{code2video}}
The baseline follows a simplified Planner $\rightarrow$ Coder $\rightarrow$ Critic pipeline inspired by Code2Video-style educational video generators. Given the final scientific figure $F$ and the gold narration transcript $\{s_i^\ast\}_{i=1}^{T}$, the planner first decomposes the narration into temporally ordered lecture chunks and assigns each chunk to a coarse spatial anchor region (e.g., LEFT, RIGHT, TOP, CENTER, or BOTTOM). These anchors are selected heuristically from narration cues and optionally refined using an LLM-based storyboard planner.

The coder stage then generates executable Manim code that renders the original figure as a persistent background image while overlaying animated bounding rectangles, subtitles, step tags, and coarse highlight regions corresponding to the assigned anchors. The renderer synchronizes highlight durations with narration timing to produce a narrated walkthrough video. Unlike \sysname{}, which grounds narration to semantically extracted figure regions and interaction structures, this baseline grounds only to coarse spatial regions defined over a fixed grid layout.

A lightweight critic/refinement stage retries rendering with safer subtitle placement and reduced overlap when rendering errors occur. However, the system does not perform OCR-based decomposition, semantic clustering, narration-aware packet construction, or validator-based grounding repair. In particular, there is no equivalent of the Perception Team $\Phi$ or the constrained grounding inventory $U$ used by \sysname{}.

\paragraph{Implementation Details of TheoremExplainAgent \citep{ku-etal-2025-theoremexplainagent}}
The baseline follows a simplified Figure Summary $\rightarrow$ Storyboard Planning $\rightarrow$ Manim Rendering pipeline inspired by TheoremExplainAgent-style educational video generators. Given the final scientific figure $F$ and the gold narration transcript $\{s_i^\ast\}_{i=1}^{T}$, the system first constructs a structured figure summary using a multimodal LLM. The model receives both the figure image and narration and produces a JSON representation containing a coarse figure summary together with drawable visual elements such as boxes, text regions, and arrows. Each extracted element contains a semantic label and a normalized spatial region over the figure.

Using this figure summary and narration, a planning module generates a storyboard consisting of temporally ordered visual actions. Each narration chunk is associated with an action type (e.g., \texttt{highlight}, \texttt{zoom}, \texttt{callout}, or \texttt{summary}) and a corresponding coarse spatial region over the figure. An LLM-based planner optionally refines the action sequence by aligning narration chunks with the extracted drawable elements and generating normalized highlight regions grounded in the summarized figure structure.
The renderer then generates executable Manim code that keeps the original scientific figure as a persistent background image while overlaying animated highlight boxes, zoom effects, callouts, subtitles, and narration-synchronized transitions. Unlike \sysname{}, which grounds narration to semantically validated region inventories and interaction structures, this baseline operates over coarse drawable figure summaries generated by the multimodal planner. The resulting highlights therefore correspond to approximate schematic regions rather than explicitly validated semantic figure components.
A lightweight retry-based refinement stage re-renders the video in safer layout configurations when rendering failures occur. However, the system does not perform OCR-grounded decomposition, semantic clustering, narration-aware packet construction, constrained region selection, or validator-based grounding repair. In particular, there is no equivalent of the Perception Team $\Phi$ or grounded inventory $U$ used by \sysname{}.

\end{document}